%% file: main.tex
\documentclass[runningheads]{llncs}
\usepackage[T1]{fontenc}
\usepackage[pdftex]{graphicx}

\usepackage[misc]{ifsym}
\usepackage{orcidlink}

\usepackage[colorinlistoftodos,prependcaption,textsize=tiny]{todonotes}

\usepackage{wrapfig}

\input{packages}

\toctitle{R\'{e}nyi Divergence Deep Mutual Learning}
\tocauthor{Weipeng~Fuzzy~Huang,Junjie~Tao,Changbo~Deng,Ming~Fan,Wenqiang~Wan,Qi~Xiong,Guangyuan~Piao}

\begin{document}
\title{R\'{e}nyi Divergence Deep Mutual Learning}
\titlerunning{R\'{e}nyi Divergence Deep Mutual Learning}
%
\author{Weipeng~Fuzzy~Huang\inst{1}\orcidlink{0000-0003-4620-6912} \Letter
\and
Junjie~Tao\inst{2} \and
Changbo~Deng\inst{1} \and
Ming~Fan\inst{3} \and
Wenqiang~Wan\inst{1} \and
Qi~Xiong\inst{1} \and
Guangyuan~Piao\inst{4}
}
\authorrunning{W.F. Huang et al.}
%
\institute{
Tencent Security Big Data Lab, Shenzhen, China \\
\email{\{fuzzyhuang,changbodeng,johnnywan,keonxiong\}@tencent.com}
\and
School of Software Engineering, Xi'an Jiaotong University, Xi'an, China \\
\email{taojunjie@stu.xjtu.edu.cn}
\and
The MoEKLINNS Laboratory, School of Cyber Science and Engineering, Xi’an Jiaotong University, Xi’an 710049, China\\
\email{mingfan@mail.xjtu.edu.cn}
\and
Department of Computer Science, Maynooth University, Maynooth, Ireland \\
\email{guangyuan.piao@mu.ie}
}

\maketitle              

\input{abstract}
\input{intro}
\input{approach}
\input{experiments}
\input{relatedwork}
\input{conclusion}

\subsection*{Acknowledgements}
We are grateful to the anonymous reviewers for their insightful feedback that has helped improve the paper.We also thank Jinghui Lu for guiding us toward this intriguing topic.

Junjie Tao and Ming Fan were supported by National Key R\&D Program of China (2022YFB2703500), National Natural Science Foundation of China (62232014, 62272377), and Young Talent Fund of Association for Science and Technology in Shaanxi, China.

\subsection*{Ethical Statement}
This paper investigates a Deep Learning optimization paradigm using publicly available datasets from the web.
As such, we assert that there are no ethical concerns in this study.

%
%
%
\bibliographystyle{splncs04}
\bibliography{main}

\end{document}


\maketitle

\input{appendix}

\bibliographystyle{acm}
\bibliography{main}

%% file: packages.tex
\usepackage{booktabs}
\usepackage{subfig}
\usepackage{caption}
\usepackage{hyperref}       
\hypersetup{colorlinks,
            linkcolor=blue,
            citecolor=black,
            urlcolor=blue,
            linktocpage,
            plainpages=false}

\usepackage{nicefrac}       
\usepackage{microtype}      
\usepackage{upgreek}
\usepackage[T1]{fontenc}
\usepackage{lmodern}

\usepackage[vlined, linesnumbered]{algorithm2e}

\usepackage{adjustbox}

\usepackage[most]{tcolorbox}
\usepackage{appendix}
\usepackage{bbm}
\usepackage{bm}
\usepackage{enumitem}
\usepackage{dsfont}
\usepackage{amsmath}
\usepackage{amssymb}
\usepackage{mathtools}
\usepackage[capitalise]{cleveref}
\usepackage{multirow}
\usepackage{url}
\usepackage{xurl}

\usepackage{enumitem}

\newtheorem{assumption}{Assumption}

\crefname{definition}{Definition}{Definitions}
\Crefname{definition}{Definition}{Definitions}
\crefname{remark}{Remark}{Remarks}
\Crefname{remark}{Remark}{Remarks}
\crefname{assumption}{Assumption}{Assumptions}
\Crefname{assumption}{Assumption}{Assumptions}

\newcommand{\renyidiv}{D_{\alpha}}
\newcommand{\renyi}{R\'{e}nyi}
\newcommand{\rdml}{RDML}

\makeatletter
\newcommand*{\rom}[1]{\expandafter\@slowromancap\romannumeral #1@}
\makeatother

\DeclareMathOperator{\Cat}{\mathrm{Categorical}}

\DeclareMathOperator{\obj}{\mathcal{L}}

\DeclareMathOperator{\E}{\mathbb{E}}

\newcommand{\m}[1]{\mathbf{#1}}

\newcommand{\X}{\m{x}}
\newcommand{\data}{\mathcal{D}}

\newlist{RQ}{enumerate}{1}
\setlist[RQ]{label=\quad RQ\arabic*:, leftmargin=35pt}

%% file: abstract.tex

\begin{abstract}
This paper revisits Deep Mutual Learning (DML), a simple yet effective computing paradigm.
We propose using \renyi{} divergence instead of the Kullback–Leibler divergence, which is more flexible and tunable, to improve vanilla DML.
This modification is able to consistently improve performance over vanilla DML with limited additional complexity.
The convergence properties of the proposed paradigm are analyzed theoretically, and Stochastic Gradient Descent with a constant learning rate is shown to converge with $\mathcal{O}(1)$-bias in the worst case scenario for nonconvex optimization tasks.
That is, learning will reach nearby local optima but continue searching within a bounded scope, which may help mitigate overfitting.
Finally, our extensive empirical results demonstrate the advantage of combining DML and the~\renyi{} divergence, leading to further improvement in model generalization.
\end{abstract}

%% file: intro.tex

\section{Introduction}
An appealing quality of certain machine learning approaches is their strong association with realworld phenomena.
Among these techniques is Deep Mutual Learning (DML)~\cite{zhang2018deep}, an empirically powerful paradigm that is, despite its conceptual simplicity, highly effective in practice.
The metaphor for DML can be likened to a learning process in which a group of students acquires knowledge not only from the ground truth but also from their peers.
Intuitively, this learning paradigm fosters the development of students by encouraging them to assimilate the strengths of one another.
As a result, the students' performance surpasses what could have been achieved if they solely relied on their teacher for guidance.
Empirically, DML is remarkable in constraining the generalization errors and hence nicely protects the learned models from overfitting by incorporating the Kullback–Leibler (KL) divergence between peers into its loss function~\cite{park2020diversified}.
Heuristically, it helps the students find wider local optima~\cite{zhang2018deep}, since the students share diversity with others and avoid being optimized towards a limited number of directions.
In fact, DML is an efficient paradigm, as it adheres to the principle of Occam's razor, where simplicity is preferred over complexity.

Interestingly, even if every student uses the same network architecture (but with distinct weight initializations), each individual model still benefits from the paradigm and outperforms itself~\cite{zhang2018deep}.
It is reasonable as weight initialization plays a crucial role in Deep Learning (DL) optimizations because some initial points can lead to wider optima than others, in the highly nonconvex optimization tasks~\cite{erhan2010does}.
To facilitate DML for the pretrained models, one may apply the \textsc{Dropout}~\cite{Srivastava2014} to produce a set of diversified initializations.

Observing that the empirical advantages of DML originate from the regularization aspect wherein students learn from others, we propose that increasing the flexibility of adjusting the regularization can further improve learning performance.
The proposed paradigm is named~\renyi{} Divergence Deep Mutual Learning (\rdml{}), which utilizes the~\renyi{} divergence~\cite{Erven2014} to regulate the degree to which a student should learn from others.
\rdml{} is simple in the same order of magnitude as DML, but performs consistently better in practice.
As a super class of the KL divergence, the~\renyi{} divergence introduces more flexibility for tuning.
Analogous to other regularization approaches, model performance benefits from a better-tuned regularization power, i.e., the coefficient for the regularization part.
The experimental study shows that~\rdml{} consistently improves DML performance.
On the theoretical side, we prove that the expected gradient norm of \rdml{} using Stochastic Gradient Descent (SGD) converges in $\mathcal{O}(1 / {\sqrt{T}} + 1)$ for every student, with a constant learning rate $o(1/{\sqrt{T}})$, where $T$ is the total number of iterations.
The reasonable amount of bias keeps the algorithm randomly searching around the (local) optima of the base model loss where every student arrives.
This could effectively increase the chance of avoiding a narrow optima in practice.

Apart from the proposal of~\rdml{}, the contributions of this paper include:
1) a theoretical analysis of the convergence properties that \rdml{} maintains;
2) an extensive empirical study in Section~\ref{sec:empirical} which shows that \rdml{} is able to consistently improve the model performance and achieve better generalizations in Computer Vision and Natural Language Processing.
Finally, the code is available at \url{http://github.com/parklize/rdml}.

%% file: approach.tex

\section{Deep Mutual Learning}
\label{sec:dml}
Before explaining DML, we introduce the common notations that are consistent throughout the paper.
We denote the instance domain set as $\mathcal{X}$ and the ground truth domain set as $\mathcal{Y}$.
Let us define the $N$-sized dataset by $\data = \{(x_n, y_n)\}_{1 \le n \le N}$, where $x_n \in \mathcal{X}$ is the $n$-th observation and $y_n \in \mathcal{Y}$ is the corresponding ground truth, for all $n$.
We also call an element $d \in \data$ a data point such that $d = (x, y)$.
Additionally, we write $\X = \{x_n\}_{1 \le n \le N}$ and $\m{y} = \{y_n\}_{1 \le n \le N}$.
Let $\renyidiv(\cdot)$ denote the~\renyi{} divergence parameterized with $\alpha$.
Moreover, we denote the indices of the students/models in the DML paradigm by $\m{s} = \{1, \dots, K\}$.
Finally, let $\eta_t$ denote the learning rate in the optimization techniques at time $t$.

A neural network can be considered a blackbox function that approximates a hard-to-define distribution.
Let $\bm{\theta} \in \Theta \subseteq \mathbb{R}^h$ be the parameter set in the neural network.
Apart from that, let us define the variables for the model by $\bm{\mu} = \{ \mu_m \}_{1 \le m \le M}$.
Taking multiclass classification as an example, we follow Kingma and Welling~\cite{kingma2019introduction} to formulate
\begin{align*}
\mathbb{P}_{\bm{\theta}} = \mathrm{NeuralNet}(x; \bm{\theta}) \qquad
p(y | x, \bm{\theta}) = \Cat(y; \mathbb{P}_{\bm{\theta}})
\end{align*}
where $\mathbb{P}_{\bm{\theta}}=\{ p(\mu | \X, \bm{\theta}) : \mu \in \bm{\mu} \}$ and a {\sc SoftMax} layer is applied to ensure $\sum_{\mu} p(\mu | x, \bm{\theta}) = 1$. We call $p(\mu | \cdot)$ the base model as it represents the applied neural network.
Next, we will introduce the \rdml{} framework before analyzing its properties.

\subsection{R\'{e}nyi Divergence Deep Mutual Learning}
Imagine that there is a cohort of students learning a task together.
The individual DML loss of each student $k$ is denoted by $\obj_k$.
Specifically, $\obj_k \coloneqq \obj_k^{base} + \obj_k^{div}$
where $\obj^{base}_k$ is the loss of the base model  for the input data and ground truth, while $\obj^{div}_k$ is the divergence loss from this student to others.

In general, the base loss $\obj^{base}_k$ is selected depending on the task and the data.
The base mode loss $\obj_k^{base}$ considers an additive ``negative log likelihood'' alongside the observation $\m{y}$.
For instance, in a multiclass classification task containing $M$ classes, the base loss is identical to the cross entropy loss, i.e.,
\begin{align}
\obj_k^{base}
\coloneqq - \frac 1 N \sum_{n} \sum_{m} \mathbbm{1}(y_n = m) \log p(y_n = m | x_n, \bm{\theta}_k)
\end{align}
where $p(y_n = m | x_n, \bm{\theta}_k)$ denotes the probability of $x_n$ belonging to class $m$ with regard to model $k$ and $\theta_k$ is the corresponding parameter set.
Also, we apply a vectorized binary random variable $\bm{\mu}_n = \{ \mu_{n m} \}_{1 \le m\le M}$, in which each $\mu_{n m}$ denotes the event of $y_n = m$.
We further denote its distribution by $\mathbb{P}(\bm{\mu}_n | x_n, \bm{\theta}_k)$.
Importantly, we emphasize that $p(\bm{\mu}_n | x_n, \bm{\theta}_k)$ can be specified for different students, i.e., there can be different models and parameters among the students.

The divergence loss $\obj_k^{div}$ depends on the approximated probabilities of the variables $\bm{\mu}$ for each student $k$.
Instead of utilizing the KL divergence as in DML, we propose to employ the~\renyi{} divergence, $\renyidiv(\cdot || \cdot)$, in \rdml{} for this part.
We define
\begin{align}
\label{eq:dml_div}
\obj^{div}_{k}
&\coloneqq \frac{1}{N (K-1)} \sum_{j \in \m{s}_{\neg k}} \sum_{n} \renyidiv [ \mathbb{P}(\bm{\mu}_n | x_n, \bm{\theta}_{j}) || \mathbb{P}(\bm{\mu}_n | x_n, \bm{\theta}_k)]
\end{align}
where $\m{s}_{\neg k} = \m{s} \setminus \{ k \}$ is the peer set of student $k$.
It indicates that the model $k$ will be calibrated by the other models.
We again emphasize that this paradigm can be trivially extended to a diversity of machine learning tasks, as shown in DML~\cite{zhang2018deep}.
In the sequel, we will discuss the~\renyi{} divergence and its usage in~\rdml{}.

\subsubsection{\renyi{} Divergence}
With a controlling parameter $\alpha \in [0, 1) \cup (1, \infty)$, the divergence is a statistical distance quantity which measures the distance from distribution $\mathbb{Q}$ to $\mathbb{P}$, defined by
\begin{align}
\label{eq:renyi}
\renyidiv(\mathbb{P} || \mathbb{\mathbb{Q}})
&= \frac 1 {\alpha-1} \log \int p(\mu)^\alpha q(\mu)^{1-\alpha} d \mu \, .
\end{align}
Same as the KL divergence, the~\renyi{} divergence is asymmetric and hence generally not a metric~\cite{Erven2014}.
It is worth noting that the~\renyi{} divergence covers a family of statistical distances.
For instance, $\alpha=0.5$ leads the~\renyi{} divergence to the squared Hellinger divergence and $\alpha \to 1$ leads to the KL divergence, etc.~\cite{li2016renyi}.
Therefore, the vanilla DML with the KL divergence can be seen as a special case of RDML.
Furthermore, the following essential remarks have been proved~\cite{Erven2014}.
\begin{remark}
\label{rmk:non-negative}
For any distribution $\mathbb{P}$, $\mathbb{\mathbb{Q}}$, and $\alpha \in [0, 1) \cup (1, \infty)$, the~\renyi{} divergence $\renyidiv(\mathbb{P} || \mathbb{\mathbb{Q}}) \ge 0$.
The equality holds if and only if $\mathbb{P}$ is identical to $\mathbb{\mathbb{Q}}$.
\end{remark}
\begin{remark}
\label{rmk:increasing}
For any distribution $\mathbb{P}$, $\mathbb{\mathbb{Q}}$, and $\alpha \in [0, 1) \cup (1, \infty)$, the~\renyi{} divergence $\renyidiv(\mathbb{P} || \mathbb{\mathbb{Q}})$ is nondecreasing in $\alpha$.
\end{remark}
The first remark fixes the lower bound of the divergence and ensures $\obj_k^{div}$ to be nonnegative, which is always a desired property for the loss function.
\cref{rmk:increasing} implies that $\alpha$ influences the distance value scope when $\mathbb{P}$ and $\mathbb{\mathbb{Q}}$ are fixed.
In the context of RDML, a larger $\alpha$ value hence pushes the students to learn more from their peers considering that the gradients for updating the parameters become numerically larger.
\begin{figure}[!t]
\centering
\subfloat{
	\includegraphics[width=0.3\textwidth]{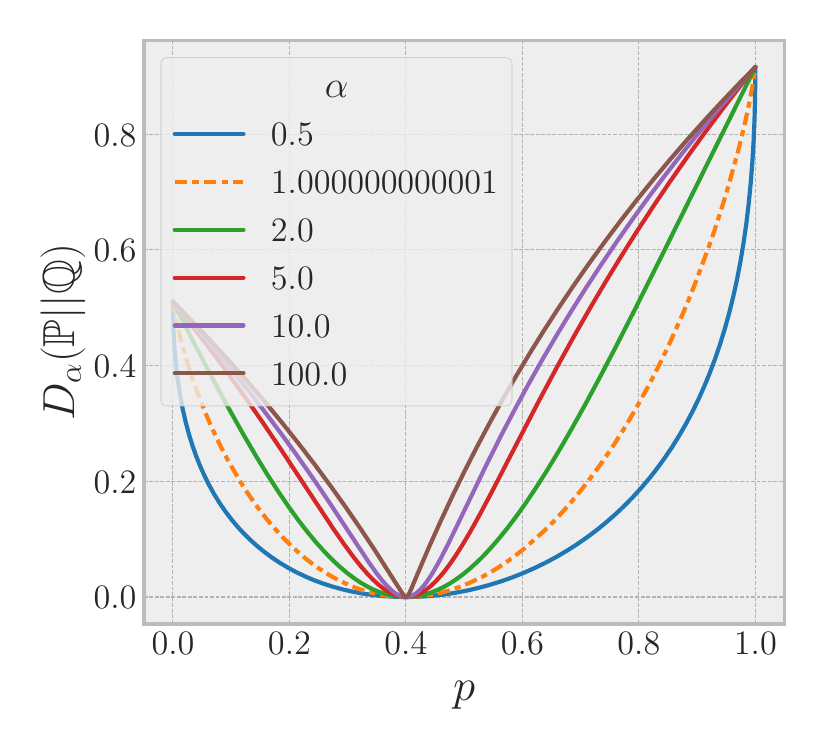}
}
\quad \quad \quad
\subfloat{
	\includegraphics[width=0.3\textwidth]{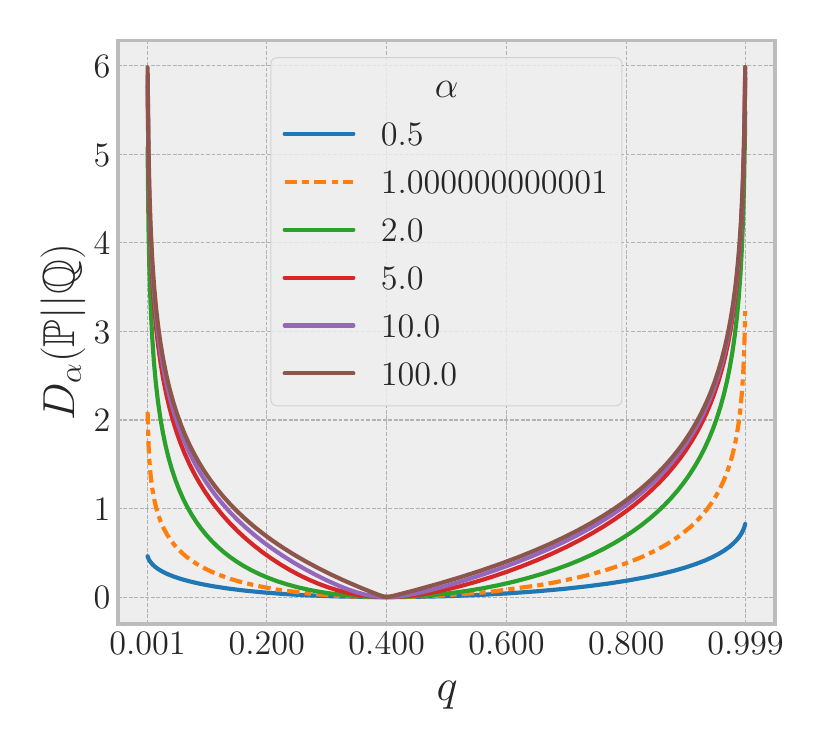}
}
\caption{
Example plots of~\renyi{} divergence for distributions containing two events.
The orange dashed line represents the KL divergence in both plots.
The first plot fixes distribution $\mathbb{Q}$ to $(0.4, 0.6)$ and shows the divergence change over $p$ and $1-p$.
The second plot fixes $\mathbb{P}= (0.4, 0.6)$ and shows the divergence change over $q$ and $1-q$.
Note that when $q=0$ or $q=1$, the divergence value is $\infty$ for any $\alpha \in [0, 1) \cup (1, \infty)$.
As infinity is not graphable, the x-axis in the second plot ranges from $0.001$ to $0.999$.
}
\label{fg:renyi_div}
\end{figure}

\cref{fg:renyi_div} illustrates an example of the $\renyidiv(\mathbb{P}||\mathbb{Q})$ for various $\alpha$ with fixed $\mathbb{P}$ and $\mathbb{Q}$, respectively.
In the example,~\renyi{} divergence with $\alpha \to 1$ is equivalent to the KL divergence.
We observe that the correlation between $\renyidiv(\mathbb{P}||\mathbb{Q})$ and $\alpha$ embodies~\cref{rmk:increasing}.
On the other hand, the case focusing on $\renyidiv(\mathbb{P} || \mathbb{Q})$ with fixed $\mathbb{P}$, spans a much broader range.
The divergence is more sensitive for $\mathbb{P} \ne \mathbb{Q}$ in this scenario with sufficiently large $\alpha$.
That said, the gap between $\alpha = 10$ and $\alpha=100$ is smaller, indicating that the divergence growth will become slower as $\alpha$ increases.
These observations imply that $\alpha$ introduces the flexibility of controlling the degree to which students learn from others in RDML.
In this study, we consider $\alpha$ as a hyperparameter that can be determined through grid-search using a validation set.
While automatically tuning $\alpha$ would be ideal, a preliminary test in which $\alpha$ was treated as a learnable parameter based on the training data did not yield improved performance.
We leave the exploration of automatic $\alpha$ tuning to future work.

\section{Properties of RDML}
In this section, we will elaborate on the convergence properties of~\rdml{}.
Additionally, we will examine the computational complexity of the paradigm.
We leave all the detailed proof in the supplemental document (SD)\footnote{The SD is available in the full version (\url{https://arxiv.org/pdf/2209.05732.pdf})}.

\begin{algorithm}[!t]
\caption{\textsc{SGD for (R)DML}}
\label{alg:dml}
\DontPrintSemicolon
Initialize $\bm{\theta}_{1, 1}, \dots, \bm{\theta}_{K, 1}$ \;
\For{$t = 1, \dots $}{
	\For{$k = 1, \dots, K$}{
		Sample a data point $d$ from $\data$ and a peer $j$ from $\m{s}_{\neg k}$ uniformly \;
		$\bm{\theta}_{k, t+1} \gets \bm{\theta}_{k, t} - \eta_t \nabla \obj_k(d, j, \bm{\theta}_{k, t})$ \;
	}
}
\end{algorithm}
Let us denote the parameter for student $k$ at time $t$, by $\bm{\theta}_{k, t}$.
To apply SGD, we write $\obj_k (d, j, \bm{\theta}_{k, t})$ the loss function parameterized with a random data point $d \in \data$, a peer $j\in \m{s}_{\neg k}$, and the current parameter $\bm{\theta}_{k, t}$.
Following that,~\cref{alg:dml} sketches the SGD procedure.
The algorithm iterates through each student and optimizes the parameters upon the corresponding DML loss.
At each single iteration, for a student $k$, we uniformly sample a data point $d$ from $\data$ and a peer $j$ from its peers $\m{s}_{\neg k}$.

Furthermore, we denote $\obj_k(\bm{\theta}_{k, t})$ as the loss that takes the entire dataset and peers as input.
Let $\E_{d, j}[\cdot]$ be abbreviated to $\E [\cdot]$.
We claim that the expected gradients are unbiased in SGD for RDML.
\begin{proposition}
\label{prop:unbiased}
For any student $k$ at any time $t$, the expected gradient for $\obj_k$ is an unbiased estimator of the gradient, such that $\E [\nabla \obj_k(d, j, \bm{\theta}_{k, t})] = \nabla \obj_k (\bm{\theta}_{k, t}), \forall k$.
\end{proposition}
However, we emphasize that bias for learning will still be generated at each iteration as the objectives are time-varying.

\subsection{Convergence Guarantee}
This section analyzes the convergence properties of RDML using SGD.
Our theoretical statements are presented here, while all the proofs are provided in the SD.
Motivated by the fact that the majority of neural networks are highly nonconvex, we focus on the convergence guarantee for the nonconvex setting.
We follow the work of~\cite{honorio2011lipschitz,cheng2018convergence} to state the following assumption for Lipschitz continuity in the probabilistic models, which is crucial for conducting further convergence analysis.
\begin{assumption}
\label{asmp:cont}
There exists a constant $L >0$.
Given any student $k$, its base model $\mathbb{P}(\mu| x, \bm{\theta})$ is $L$-Lipschitz continuous in $\bm{\theta}$, such that $\forall \bm{\theta}, \bm{\theta}' \in \Theta$,
\begin{align*}
 | p(\mu| x, \bm{\theta}) - p(\mu| x, \bm{\theta}') |
&\le L \| \bm{\theta} - \bm{\theta}' \|
\end{align*}
for every realization of $\mu$.
Note that a function is usually not $L$-Lipschitz continuous on the unbounded domain.
Therefore, we further assume that there exists some $\tau \in (0, 1)$ satisfying $\forall \mu, x, \bm{\theta}: p(\mu | x, \bm{\theta}) \ge \tau$
throughout the entire learning process.

In addition, the base models satisfy the $H$-smoothness in $\bm{\theta}$, such that
\begin{align*}
\forall \bm{\theta}, \bm{\theta}'\in \Theta: \quad \| \nabla p(\mu| x, \bm{\theta}) - \nabla p(\mu| x, \bm{\theta}') \|
&\le H \| \bm{\theta} - \bm{\theta}' \|,
\end{align*}
for every realization\footnote{Following~\cite{nguyen2018sgd}, we use this term \emph{realization} to refer to an instance of variable in the valid domain} of $k, d, \mu$.
\end{assumption}

\begin{assumption}
\label{asmp:smooth}
For any student $k$ and data point $d \in \data$, the base loss
$\obj_k^{bass}(d, \bm{\theta})$ taking $d$ and $\bm{\theta}$ as input, is differentiable and $V$-smooth in $\bm{\theta}$ on the bounded domain, i.e.,
\begin{align*}
\forall \bm{\theta},\bm{\theta}'\in \Theta: \quad \| \nabla \obj_k^{bass}(d, \bm{\theta}) - \nabla \obj_k^{bass}(d, \bm{\theta}') \| \le V \| \bm{\theta} - \bm{\theta}' \| \,.
\end{align*}
\end{assumption}
Given the assumptions, we can set up the property of smoothness for each individual loss $\obj_k$.
Smoothness is the most fundamental condition which the convergence analysis of the nonconvex optimization should satisfy~\cite{nesterov2003introductory}.
\begin{theorem}
\label{thm:q_smooth}
Let $W = V + |\alpha-1| \varphi^2 + \alpha L^2 + \tau H $ where
$\varphi = \min \left\{ {\tau}^{-1}, {M/(\tau e^{(M-1) \tau})^{\alpha}} \right\} L$.
Suppose~\cref{asmp:smooth,asmp:cont} satisfy, the loss $\obj_k$ for every student $k$ is $W$-smooth in its parameter $\bm{\theta}_k$ provided that $\bm{\theta}_j$ is fixed for all $j \in \m{s}_{\neg k}$.
\end{theorem}
\cref{thm:q_smooth} enables us to apply the standard convergence analysis to this paradigm.
Now, we introduce another common assumption (\cref{asmp:bounded_grad}) for the convergence analysis of nonconvex problems.
\begin{assumption}
\label{asmp:bounded_grad}
For every student $k$ and time $t$, $\E [\| \nabla \obj_k^{base}(d, \bm{\theta}_{k, t}) \|^2] \le \tilde{\sigma}^2$.
\end{assumption}
This is a general assumption in nonconvex convergence analysis, but implies the following lemma bounding the noise for the whole~\rdml{} objectives along with that each $p(\cdot) \ge \tau$ in~\cref{asmp:smooth}.
\begin{lemma}
\label{lem:bounded_noise}
Based on~\cref{asmp:smooth,asmp:bounded_grad}, $\E[ \|\nabla \obj_k(d, j, \bm{\theta}_{k, t}) \|^2 ] \le \sigma^2$ for every realization of $k$ and $t$, where $\sigma^2 = 2\tilde{\sigma}^2 + 2\varphi^2$.
\end{lemma}

Next, we will show the worst case convergence of the expected gradient norm.
However, we defer the discussion of \emph{conditional} convergence in  $\mathcal{O} ( {1}/{\sqrt{T}} )$ in the average case with a constant learning rate of $\Omega ( {1}/{\sqrt{T}} )$ to the SD.
It is difficult to determine whether this condition, which leads to a unbiased convergence, can be satisfied without further information.
Thus, we claim our formal convergence result only for the worst case scenario.
\begin{theorem}
\label{thm:converge}
Let $\obj^{base}_{k ,*}$ be the global minima of $\obj^{base}_k$.
Under~\cref{asmp:cont,asmp:smooth,asmp:bounded_grad}, by selecting a constant learning rate $\eta_t = \frac{\eta}{\sqrt{T}} \le \frac{1}{W \sqrt{T}}$ that depends on the total iteration $T$, $\E[\min_t \| \nabla \obj_k(\bm{\theta}_{k, t}) \|^2]$ is bounded by
\[
\mathcal{O}\left( \frac {2 (\obj^{base}_{k, 1} - \obj^{base}_{k ,*})} {\eta \sqrt{T}} + \frac{\sigma^2}{\sqrt{T}} + 2 \max(\varphi \tilde{\sigma}, \varphi^2) \right) = \mathcal{O}\left(\frac{1}{\sqrt{T}} \right) + \mathcal{O}(1)
\]
for any student $k$ in the worst case scenario.
\end{theorem}
Unlike the conventional analysis of the biased gradient estimators for SGD, in which the bias accompanies the gradients~\cite{ajalloeian2020analysis,hu2021analysis}, the bias in~\rdml{} is introduced by the time-varying objectives of~\rdml{}.
The gradients of all students will receive less impact from the base model as time progresses, but will still keep learning from the peers.
One can imagine that the $K$ parameters arrive at certain local optima, but still attempt to ``move closer'' to the peers with a reasonable pace.
This might in practice help some models escape to wider local optima.

\subsection{Computational Complexity of~\rdml{}}
We denote the time complexity for each student $k$ at a single iteration as $\mathcal{O}(B_k)$.
Also, the time cost of the~\renyi{} divergence between $\mathbb{P}(\bm{\mu} | \cdot)$ and $\mathbb{P}(\bm{\mu} | \cdot)$, for any student $j$ and $k$, depends linearly on the size of $\bm{\mu}$, which is $\mathcal{O}(M)$ ignoring the complexity of the log and power function, etc.
For a single loop, the time cost is $\mathcal{O} ( \sum_k (B_k + M)) = \mathcal{O}( \sum_k B_k + MK)$, since each student samples only one peer to learn from in the loop.
Considering that the $S_k$ is the size for each model $k$, the space complexity is trivially $\mathcal{O}(\sum_k S_k)$.

%% file: experiments.tex

\section{Empirical Study}
\label{sec:empirical}
Our empirical study aims to investigate whether~\rdml{} can enhance model performance beyond that of the independent model and vanilla DML.
Thus, we minimize the effort for tuning the models and instead adopt certain general settings.
In summary, our study focuses on the following research questions.
\begin{enumerate}
\item
How does the algorithm converge? (\cref{sec:exp_convergence})
\item
How does \rdml{} perform compared to vanilla DML and the single model on its own? (\cref{sec:eval})
\item
Does \rdml{} generalize better? (\cref{sec:generalization})
\end{enumerate}

\subsection{Experimental Setup}
This study centers on datasets from two prominent fields: Computer Vision (CV) and Natural Language Processing (NLP).
The CV data contains \texttt{CIFAR10} and \texttt{CIFAR100}~\cite{krizhevsky2009learning} with a resolution of $32 \times 32$.
We conducted experiments on \texttt{DTD}~\cite{cimpoi14describing} and \texttt{Flowers102}~\cite{Nilsback08} with a resolution of $224 \times 224$.
The NLP data for text classification task contains \texttt{AGNews}, \texttt{Yahoo!Answers}, and \texttt{YelpReviewFull}.
All data is handled through either \textbf{torchvision} or \textbf{torchtext}.\footnote{The required public resources of software and model are available in the SD.}

Regarding \texttt{CIFAR10} and \texttt{CIFAR100}, we applied SGD with Nesterov momentum of $0.9$.
The initial learning rate was set to $0.1$, and the weight decay was set to $0.0005$.
A total of $200$ epochs were run with a batch size of $128$, and the learning rate dropped by $0.2$ every $60$th epoch.
The architectures used are GoogLeNet~\cite{szegedy2014going}, ResNet34~\cite{he2016deep}, and VGG16~\cite{simonyan2015very}.
For \texttt{DTD} and \texttt{Flowers192}, we set the learning rate to $0.005$ and the weight decay to $0.0001$ using SGD with momentum of $0.9$.
We ran 30 epochs with a batch size of $32$ and decay the learning rate by $0.1$ at epochs $(16, 22)$.
In addition, the gradients were clipped by the max norm of $5$.
For these two datasets, \textbf{timm} were used to acquire the pretrained weights for the models: InceptionV4~\cite{szegedy2017inception}, ViT~\cite{wu2020visual}, and YoloV3~\cite{redmon2018yolov3}.

For text classification, we evaluated the performance of fastText~\cite{joulin2016bag} with GloVe~\cite{pennington2014glove} (pretrained word embeddings) and CharCNN~\cite{zhang2015character} using a public implementation.
We used the default settings of the implementation with an initial learning rate of $0.5$ and set the weight decay to $0.5$ for fastText.
Then, 100 epochs were run, in which the learning rate was decayed every three epochs with a batch size of $16$.
For CharCNN, we set an initial learning rate of $0.0001$ and set the weight decay to $0.9$ for the experiments for~\texttt{Yahoo!Answers} and~\texttt{YelpReviewFull}.
For all other experiments, we set the weight decay to $0.5$ and uses the default settings of the implementation, including an initial learning rate of $0.001$.
Finally, we ran 20 epochs and decayed the learning rate every three epochs with a batch size of $128$.
Even though a larger $K$ value is preferred for potentially achieving better results, we followed~\cite{zhang2018deep} to set $K = 2$, which suffices to improve model performance in this study

\begin{figure}[!t]
\centering
\subfloat{
\includegraphics[width=0.88\textwidth]{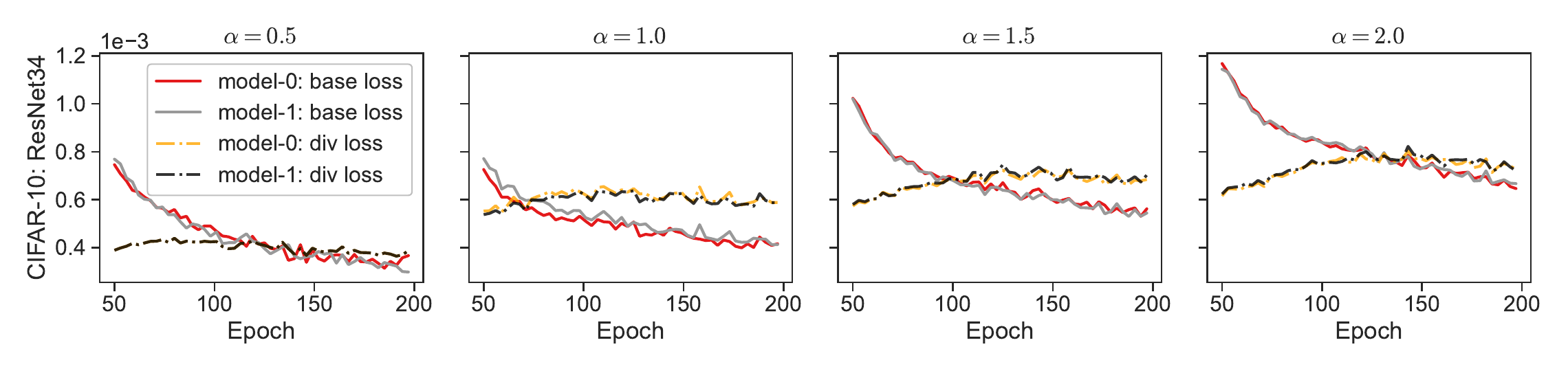}
}
\\
\subfloat{
\includegraphics[width=0.88\textwidth]{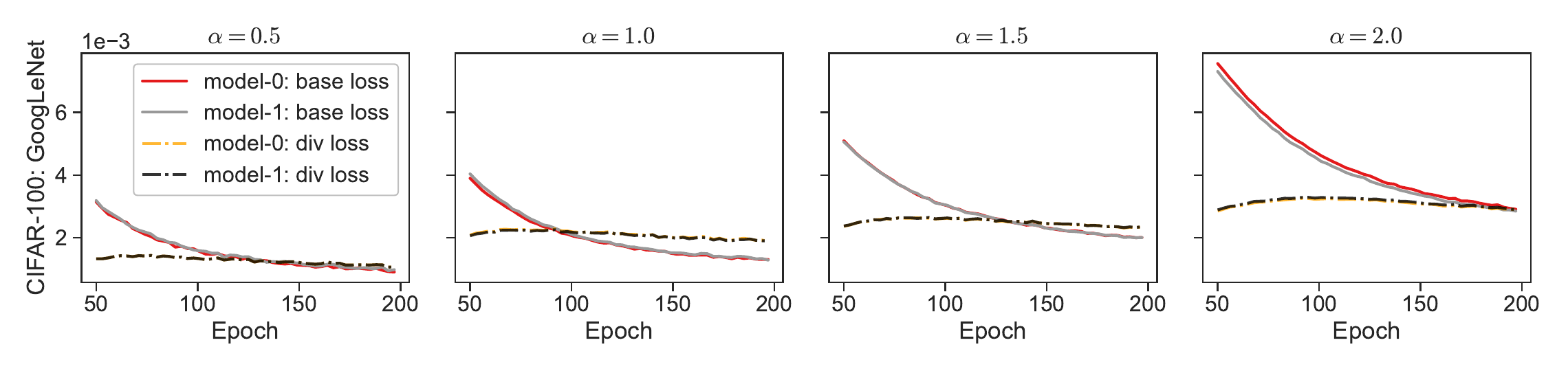}
}
\caption{
Plot of training loss for selected configurations, starting from the 50th epoch.
The base loss and divergence loss are shown separately, with values presented for every third epoch.
}
\label{fg:train_loss}
\end{figure}

\subsection{Convergence Trace Analysis}
\label{sec:exp_convergence}
We conducted a convergence analysis using the training loss trace plots depicted in~\cref{fg:train_loss}.
First, the base losses keep decreasing overall, while the divergence losses are not guaranteed to vanish but remain bounded.
This finding indirectly supports~\cref{thm:converge}, which suggests that the influence of the base loss on the gradients will reduce, but the bias introduced by the divergence loss may persist.
Secondly, we observe that the base losses can obtain lower numeric values with smaller $\alpha$ in the same time window.
Simultaneously, the divergence losses are tied to a higher level with a larger $\alpha$ value as it forces a greater learning power within the~\renyi{} divergence.
Under the selected configurations, a larger value of $\alpha$ results in a greater coefficient in the big-O notation of $\mathcal{O}(1/\sqrt{T})$, causing the error to decrease at a slower pace.
Finally, we note that for $\alpha=0.5$, the divergence losses for the two students are identical.
This is because the~\renyi{} divergence is symmetric with $\alpha=0.5$~\cite{Erven2014}, i.e., $D_{0.5}(P||Q) = D_{0.5}(Q||P)$ for any $P$ and $Q$, indicating that the divergence losses under this two-student configuration are strictly equal.

\subsection{Evaluation Results}
\label{sec:eval}
\paragraph{Image Classification.}
The training set and test set are split automatically through \textbf{torchvision}.
As for classification, we employ the top-1 accuracy to examine the model performance.
The experiments were run upon a range of configurations, including the independent case and $\alpha \in \{ 0.5, 1, 1.5, 2 \}$.
The results for \texttt{CIFAR10} and \texttt{CIFAR100} are presented in~\cref{tb:32_perf}, while those for \texttt{DTD} and \texttt{Flowers102} are displayed in~\cref{tb:224_perf}.
The indices of model are sorted by their respective values, and we label the best performer in each column using boldface.

\begin{table*}[!t]
\caption{
The top-1 accuracy ($\%$) results are presented for \texttt{CIFAR10} and \texttt{CIFAR100}.
To ensure a stable outcome, we bootstrap and average the accuracy of the test set over the last 10 epochs.
The models are listed in ascending order based on their values.
In the table, GoogLeNet, ResNet34, and VGG16 are abbreviations for GN, RN, and VG, respectively.
The model index starts from 0 and is indicated as a subscript. $\mathrm{\rdml}_{a}$ indicates that $\alpha$ is set as $a$.
The independent case (Ind.) has only one result.
The bootstrapped standard deviation is presented.
}
\label{tb:32_perf}
\centering
\begin{adjustbox}{width=1.\textwidth}
\begin{tabular}{lcccccccccccc}
\midrule
\multirow{2}{*}{} & \multicolumn{6}{c}{Dataset: \texttt{CIFAR10}} & \multicolumn{6}{c}{Dataset: \texttt{CIFAR100}} \\
\cmidrule(lr){2-7} \cmidrule(lr){8-13}
	 &  $\mathrm{GN}_{0}$ & $\mathrm{GN}_{1}$ &  $\mathrm{RN}_{0}$ & $\mathrm{RN}_{1}$ &  $\mathrm{VG}_{0}$ & $\mathrm{VG}_{1}$  &  $\mathrm{GN}_{0}$ & $\mathrm{GN}_{1}$ &  $\mathrm{RN}_{0}$ & $\mathrm{RN}_{1}$ &  $\mathrm{VG}_{0}$ & $\mathrm{VG}_{1}$ \\
\cmidrule(lr){2-7} \cmidrule(lr){8-13}
Ind.	& 	&	 $91.87\pm.14$ &          &       $92.56\pm.17$ &         &    $90.71\pm.09$ &      & $69.60\pm.19$ &          &       $76.26\pm.03$ &         &    $72.21\pm.03$       \\
$\mathrm{RDML}_{0.5}$	&	$91.91\pm.11$ &       $92.06\pm.13$ &      $92.86\pm.07$ &      $93.03\pm.07$ &    $91.03\pm.05$ &   $91.26\pm.10$ 	&        $70.22\pm.16$ &       $70.24\pm.17$ &      $76.71\pm.05$ &      $77.14\pm.02$ &    $72.09\pm.04$ &   $72.53\pm.05$ \\
$\mathrm{DML}_{}$       &	 $92.28\pm.13$ &       $92.37\pm.16$ &       $93.27\pm.08$ &      $93.34\pm.07$ &    $\underline{91.22\pm.11}$ &   $\underline{91.33\pm.07}$ &        $\underline{72.34\pm.13}$ &       $\underline{72.49\pm.18}$	&       $77.14\pm.03$		&      $77.24\pm.03$ &    $73.01\pm.05$ &   $73.17\pm.04$ \\
$\mathrm{RDML}_{1.5}$	&        $\underline{92.61\pm.11}$ &       $\underline{92.62\pm.03}$ &       $\underline{93.28\pm.06}$ &      $\underline{93.35\pm.11}$ &    $91.18\pm.11$ &  $91.29\pm.08$ &        $71.65\pm.13$ &        $72.1\pm.07$ &       $77.61\pm.03$ &      $77.94\pm.02$ &    $\underline{73.32\pm.05}$ &    $73.6\pm.06$ \\
$\mathrm{RDML}_{2.0}$	&        $92.26\pm.14$ &       $92.28\pm.12$ &      $93.18\pm.07$ &      $93.19\pm.08$ &    $91.17\pm.10$ &   $91.23\pm.07$   &        $72.13\pm.10$ &        $72.27\pm.16$ &        $\underline{78.31\pm.04}$ &        $\underline{78.5\pm.02}$ &      $73.3\pm.06$ &        $\underline{73.74\pm.03}$      \\
\bottomrule
\end{tabular}
\end{adjustbox}
\end{table*}

\begin{table*}[!t]
\centering
\caption{
The top-1 accuracy ($\%$) results for \texttt{DTD} and  \texttt{Flowers102} are presented.
The accuracy results of the last 5 epochs in the test set are bootstrapped and averaged.
In the table, IV4 and YV3 are abbreviations for InceptionV4 and YoloV3, respectively.
The independent model is denoted as Ind., and the subscript of RDML is $\alpha$.
The same models are reindexed by their values in ascending order.
The bootstrapped standard deviation is presented.
}
\label{tb:224_perf}
\begin{adjustbox}{width=1.\textwidth}
\begin{tabular}{lcccccccccccc}
\midrule
\multirow{2}{*}{ } & \multicolumn{6}{c}{Dataset: \texttt{DTD}} & \multicolumn{6}{c}{Dataset: \texttt{Flowers102}} \\
\cmidrule(lr){2-7} \cmidrule(lr){8-13}
	 &  $\mathrm{IV4}_{0}$ & $\mathrm{IV4}_{1}$ &  $\mathrm{ViT}_{0}$ & $\mathrm{ViT}_{1}$ &  $\mathrm{YV3}_{0}$ & $\mathrm{YV3}_{1}$  &  $\mathrm{IV4}_{0}$ & $\mathrm{IV4}_{1}$ &  $\mathrm{ViT}_{0}$ & $\mathrm{ViT}_{1}$ &  $\mathrm{YV3}_{0}$ & $\mathrm{YV3}_{1}$ \\
\cmidrule(lr){2-7} \cmidrule(lr){8-13}
Ind.	 &       &    $64.66\pm.20$ &            &  $72.52\pm.06$ &        &      $68.95\pm.06$ &               &        $88.32\pm.12$ &              &  $98.15\pm.00$ &        &    $90.67\pm.03$      \\
$\mathrm{RDML}_{0.5}$	&          $61.63\pm.13$ &          $64.93\pm.10$ &  $72.73\pm.02$ &  $73.87\pm.04$ &      $66.07\pm.06$ &    $67.02\pm.03$	&         $88.01\pm.12$ &       $\underline{89.39\pm.10}$ &    $98.19\pm.00$ &   $98.52\pm.01$ &  $90.66\pm.02$ &   $90.68\pm.01$ \\
$\mathrm{DML}_{}$       &	 $ 65.87\pm.19$ &           $66.20\pm.09$ &  $74.38\pm.02$ &  $75.05\pm.04$ &     $69.93\pm.07$ &     $70.24\pm.09$ &        $87.92\pm.10$ &       $88.46\pm.08$ &   $98.53\pm.01$ &  $98.73\pm.01$ &   $90.98\pm.03$ &   $91.04\pm.03$ \\
$\mathrm{RDML}_{1.5}$	&           $67.57\pm.08$ &          $67.67\pm.13$ &  $\underline{75.86\pm.04}$ &  $75.94\pm.04$ &      $70.91\pm.10$ &     $\underline{71.72\pm.07}$  &         $\underline{88.27\pm.06}$ &        $88.89\pm.11$ &  $98.73\pm.01$ &  $98.91\pm.00$ &   $\underline{91.29\pm.03}$ &   $\underline{91.37\pm.03}$ \\
$\mathrm{RDML}_{2.0}$&           $\underline{67.86\pm.16}$ &          $\underline{68.97\pm.10}$ &  $75.37\pm.05$ &  $\underline{76.04\pm.02}$ &      $\underline{71.17\pm.08}$ &     $71.26\pm.09$	&                $87.98\pm.06$ &   $88.13\pm.07$ &  $\underline{98.95\pm.00}$ &  $\underline{99.02\pm.00}$ &    $90.56\pm.03$ &   $90.79\pm.02$     \\
\bottomrule
\end{tabular}
\end{adjustbox}
\end{table*}
The results of \texttt{CIFAR10} show that for GoogLeNet and ResNet34, \rdml{} with $\alpha=1.5$ performs the best, while vanilla DML obtains the best accuracy with VGG16.
In the experiments for \texttt{CIFAR100}, the three selected architectures achieve the best outcomes with $\alpha=2$ in most cases.
Observed from the results, \rdml{} under certain configuration always outperforms the independent case.
It shows that tuning $\alpha$ is helpful in learning better model parameters.
Regarding \texttt{DTD}, $\alpha=2$ and $\alpha=1.5$ are still the best options for the selected architectures.
Same as DML, \rdml{} also perfectly collaborates with the pretrained models which are agreed to be more powerful in modern DL tasks~\cite{erhan2010does}.
Also, we notice that the improvements over the independent model are greater with these models.
The effectiveness of tuning $\alpha$ is further confirmed by the results of \texttt{Flowers102} in~\cref{tb:224_perf}.
For the three models (InceptionV4, ViT, and YoloV3), the best performance is achieved with \rdml{} when $\alpha=0.5, 2.0$, and $1.5$, respectively.
For instance, ViT with RDML\textsubscript{2.0} achieves a top-1 accuracy of 99.02\% which outperforms independent model (98.15\%) and vanilla DML (98.73\%).

\begin{table*}[!t]
\centering
\scriptsize
\caption{
The results for Top-1 accuracy ($\%$) on the test sets of \texttt{AGNews}, \texttt{Yahoo!Answers}, and \texttt{YelpReviewFull}.
The best performing $\alpha$ of RDML is selected based on the mean accuracy of $K$ models on each validation set.
}
\label{tb:nlp_perf}
\begin{tabular}{lcccccccc}
\midrule
\multirow{2}{*}{Dataset} & \multicolumn{4}{c}{fastText} & \multicolumn{4}{c}{CharCNN} \\ \cmidrule(lr){2-5} \cmidrule(lr){6-9}
	 &  Ind. & $\mathrm{fastText}_0$ &  $\mathrm{fastText}_1$ &  $\alpha$ &  Ind. & $\mathrm{CharCNN}_0$ & $\mathrm{CharCNN}_1$ & $\alpha$ \\
 \midrule
\texttt{AGNews}		&	89.72	&	\underline{89.76}	&	\underline{89.76}	& 2.0	& 87.13	& \underline{89.63}	& 89.17	& 1.5 \\
\texttt{Yahoo!Answers}	&	65.65	&	\underline{65.65}	&	\underline{65.65}	& 0.5	& 63.54	& \underline{64.54}	& 64.51	& 2.0 \\
\texttt{YelpReviewFull}	&	54.26	&	\underline{54.31}	&	\underline{54.31}	& 1.5	& 55.01	& 55.36	& \underline{55.58}	& 1.5	\\
\bottomrule
\end{tabular}
\end{table*}

\paragraph{Text Classification.}
The evaluation adopts top-1 accuracy, which is the same as that used for image classification.
We demonstrate that $\alpha$ can be tuned as a hyperparameter using a grid search on a validation set.
To this end, we chose $\alpha$ based on a grid search $\in \{0.5, 1, 1.5, 2\}$ using a validation set that makes up 20\% of the training set, and investigate the best performing $\alpha$ values and the corresponding performance on the test set.
\cref{tb:nlp_perf} presents the values of $\alpha$ that yield the highest average accuracy on the test set.
Regularization via~\rdml{} was shown to be more beneficial for the larger models in regard to improving the performance over a single model, e.g., CharCNN (with $\sim$2M parameters) versus fastText (with $\sim$3K parameters).

\begin{figure}[b]
\centering
\includegraphics[width=0.65\textwidth, trim=2.8cm 0cm 1.4cm 0cm, clip=true]{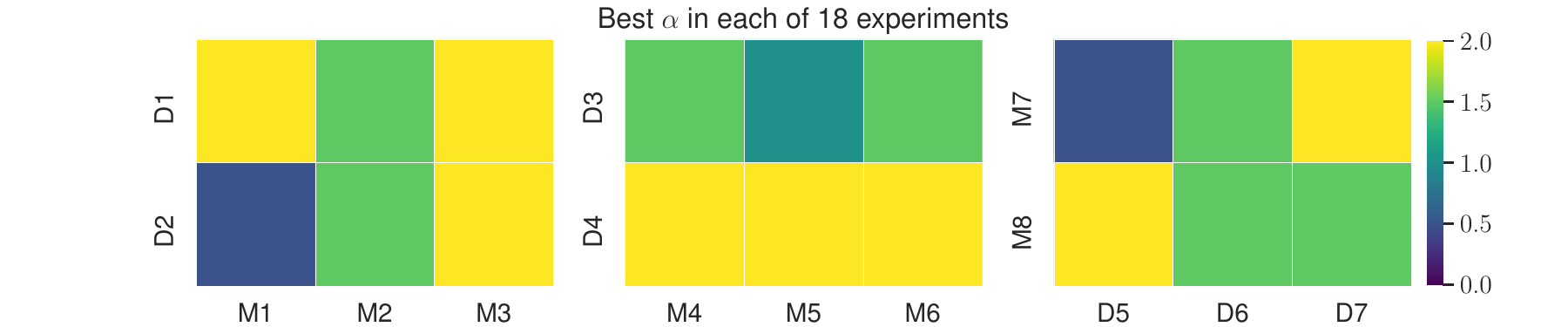}
\caption{
Heatmap depicting the optimal $\alpha$ values obtained from $18$ experiments, each corresponding to a distinct pairing of a model M and a dataset D.
The figure highlights the variability in $\alpha$ requirements for achieving superior performance when using different datasets with the same method (and vice versa).
}
\label{fg:heatmap}
\end{figure}

\paragraph{Summary.}
Observing the experiments, it is evident that \rdml{} is able to improve on vanilla DML and independent models in most cases.
This covers a broad collection of network architectures and datasets.
The best performing value of $\alpha$ varies depending on the situation as illustrated in~\cref{fg:heatmap}.
In our experiments, we found that~\rdml{} was more effective at coping with underperforming configurations, such as using a less well-defined model or a set of unoptimized hyperparameters.
This is reasonable because a perfect configuration for a moderate task suffices to obtain maximum performance.

\subsection{Generalization Results}
\label{sec:generalization}
Here, we examine the generalization ability of \rdml{} regarding the choice of $\alpha$.
We focus on the test performance of \texttt{CIFAR100} and \texttt{Flowers102}, while the other results are included in the SD.

\begin{figure}[!t]
\centering
\subfloat{
\includegraphics[width=0.74\textwidth]{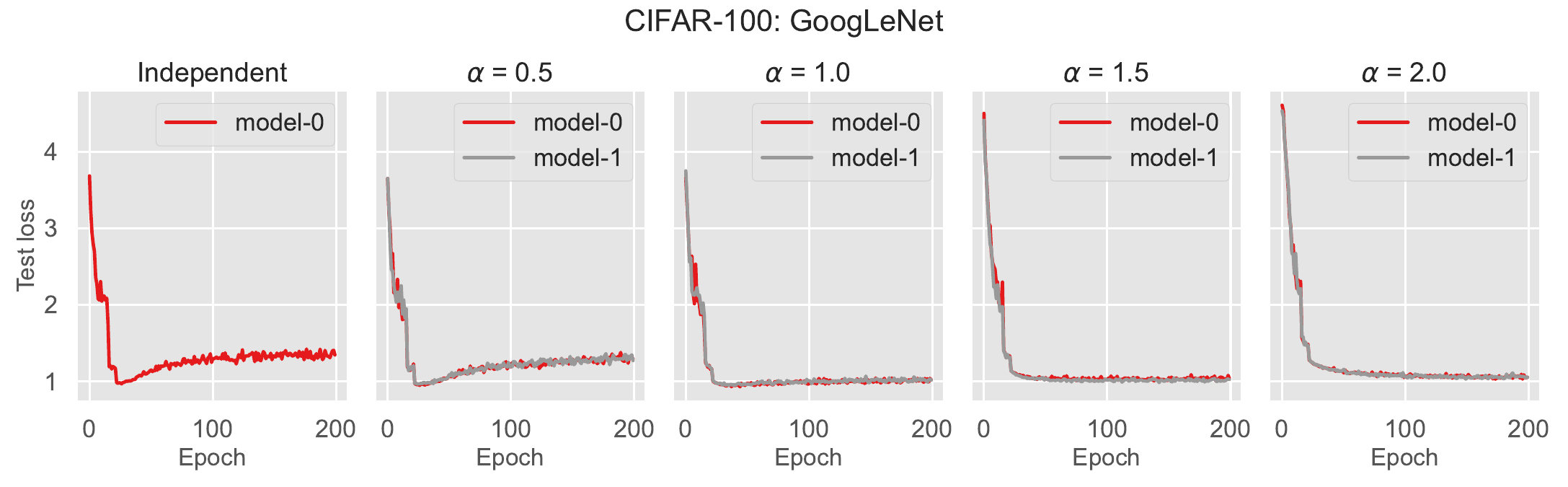}
}
\\
\subfloat{
\includegraphics[width=0.74\textwidth]{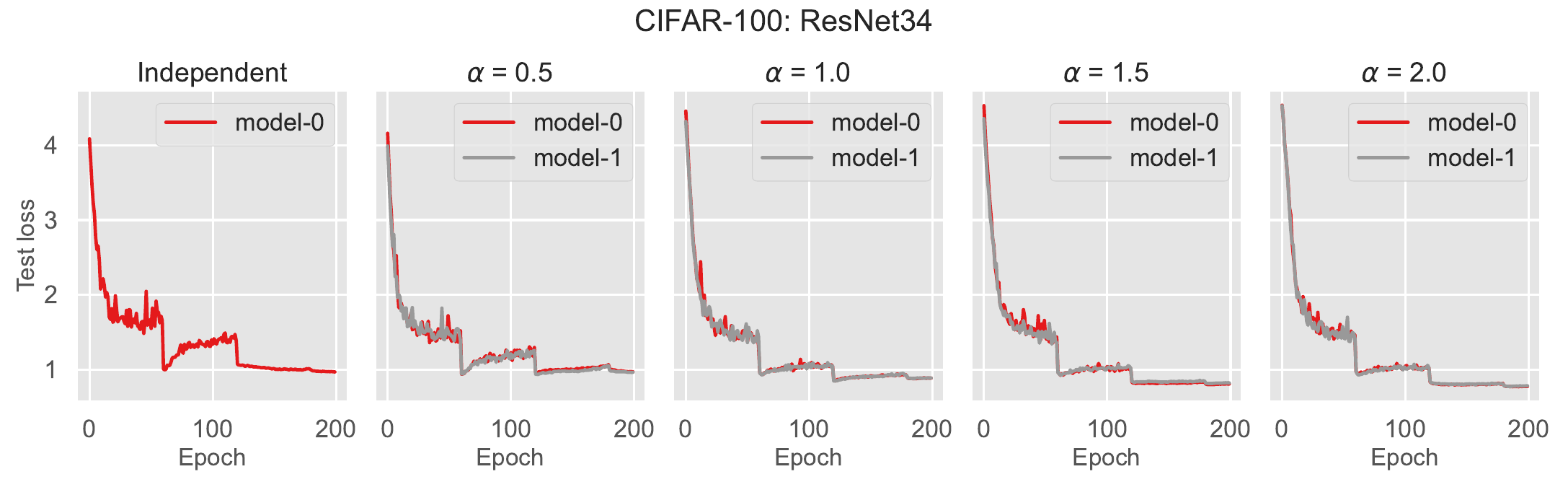}
}
\\
\subfloat{
\includegraphics[width=0.74\textwidth]{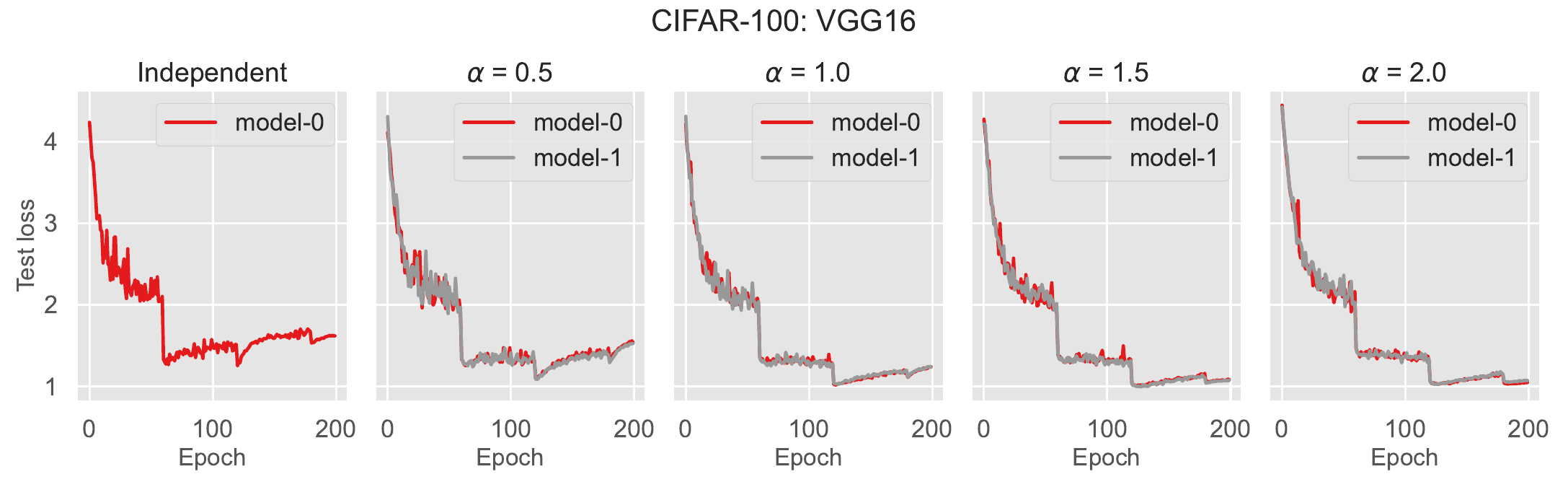}
}
\caption{Test loss of \rdml{} for various $\alpha$ on the \texttt{CIFAR100} dataset.}
\label{fg:cifar_val_loss}
\end{figure}

\begin{figure}[!t]
\centering
\subfloat{
\includegraphics[width=0.74\textwidth]{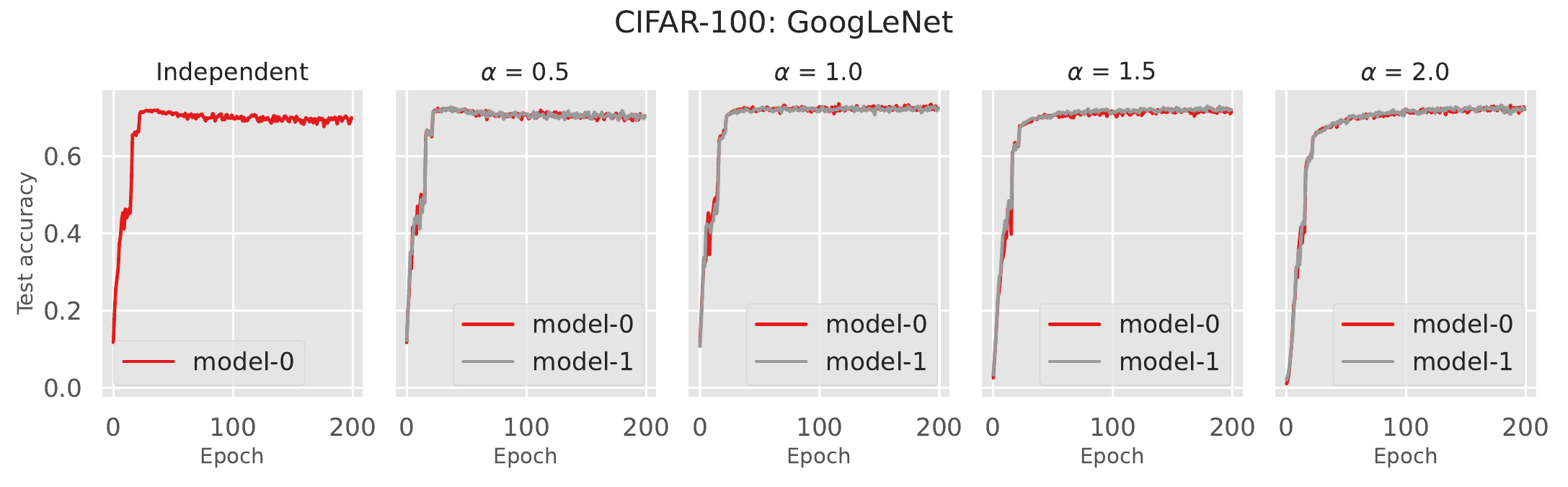}
}
\\
\subfloat{
\includegraphics[width=0.74\textwidth]{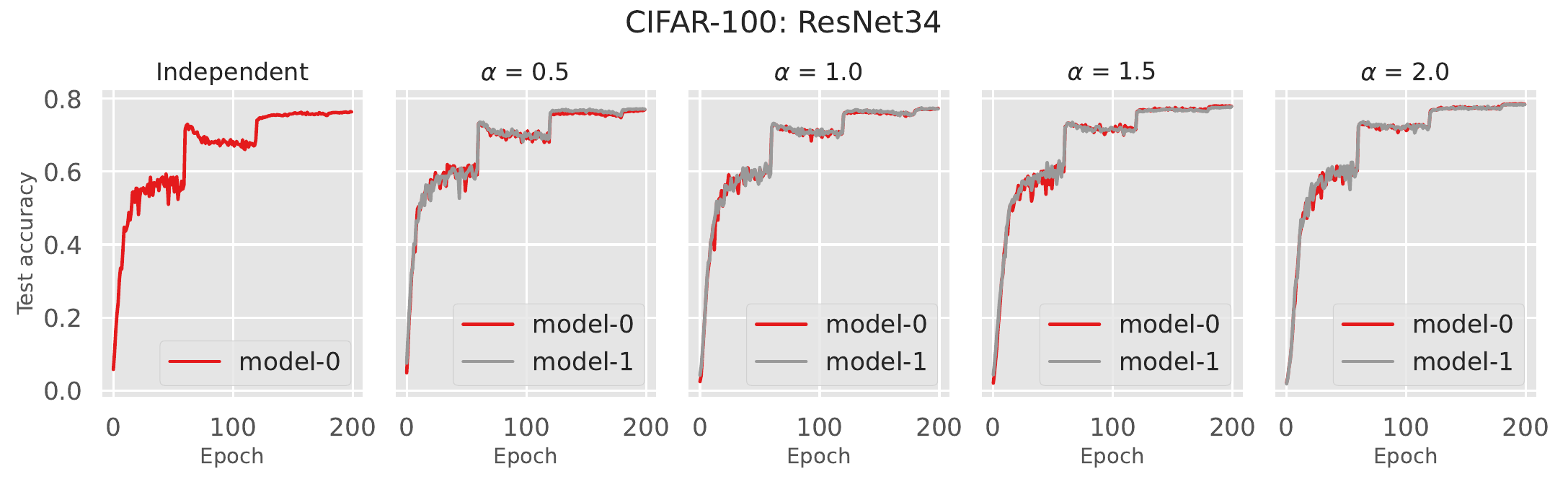}
}
\\
\subfloat{
\includegraphics[width=0.74\textwidth]{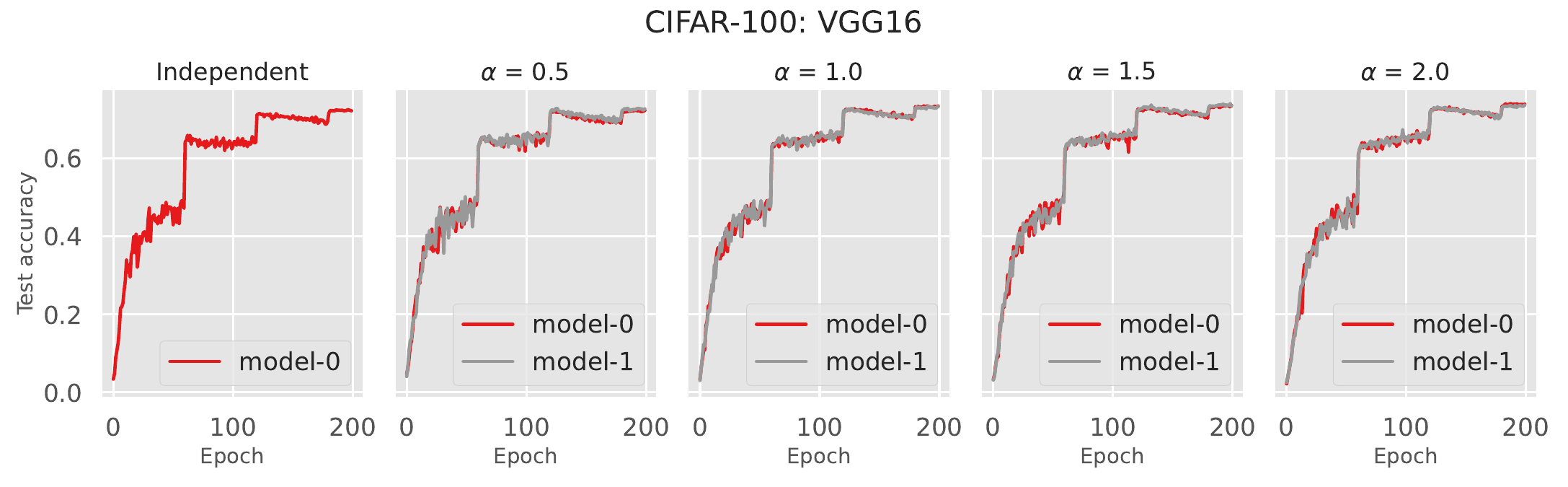}
}
\caption{Test accuracy of \rdml{} for various $\alpha$ on the \texttt{CIFAR100} dataset.}
\label{fg:cifar_val_acc}
\end{figure}
To begin,~\cref{fg:cifar_val_loss} illustrates the evaluation of the base loss on the test data using the \texttt{CIFAR100} dataset. It is widely acknowledged in the machine learning community that, beyond a certain point, while the training loss can be continuously reduced, the test loss will typically start to increase.
As demonstrated in~\cref{fg:cifar_val_loss}, the test loss of a single model (either GoogLeNet, ResNet34, or VGG16) replicates this agreement.
However, \rdml{} is able to constrain this growth tendency of test loss for all students.
Also, the capability of preventing this increase gets enhanced as $\alpha$ becomes larger, i.e., setting a prior with smaller variance.
Moreover, we observe that \rdml{} with larger $\alpha$ significantly reduces the fluctuations in the test loss, i.e., the variance is being reduced for the test loss (though it still maintains a reasonable amount of uncertainty).
It depicts that $\alpha=2$ for GoogLeNet over-regularizes the model and thus performs slightly worse than that with $\alpha=1.5$.
In the classification problem, the test base loss approximates the negative likelihood of the unseen data which can be optimized via \rdml{}.
The results suggest that \rdml{} is able to acquire better generalizations through tuning the parameter $\alpha$.

Similar patterns are observed while checking the test accuracy in~\cref{fg:cifar_val_acc}.
In regard to ResNet34, starting from epoch 60, the single model boosts its accuracy since the learning rate is decayed, but rapidly encounters a decrease in accuracy.
However, \rdml{} can significantly mitigate the performance decrease and achieve the best outcome when $\alpha=2$ within the range of our choices.
Correspondingly, the results for ResNet34 in~\cref{fg:cifar_val_loss} show a loss increase during the same period.
As shown, \rdml{} can limit this loss increasing speed and can limit harder with a greater value of $\alpha$.

\begin{figure}[!t]
\centering
\subfloat{
\includegraphics[width=0.74\textwidth, trim=0.3cm 0cm 0cm 0cm, clip=true]{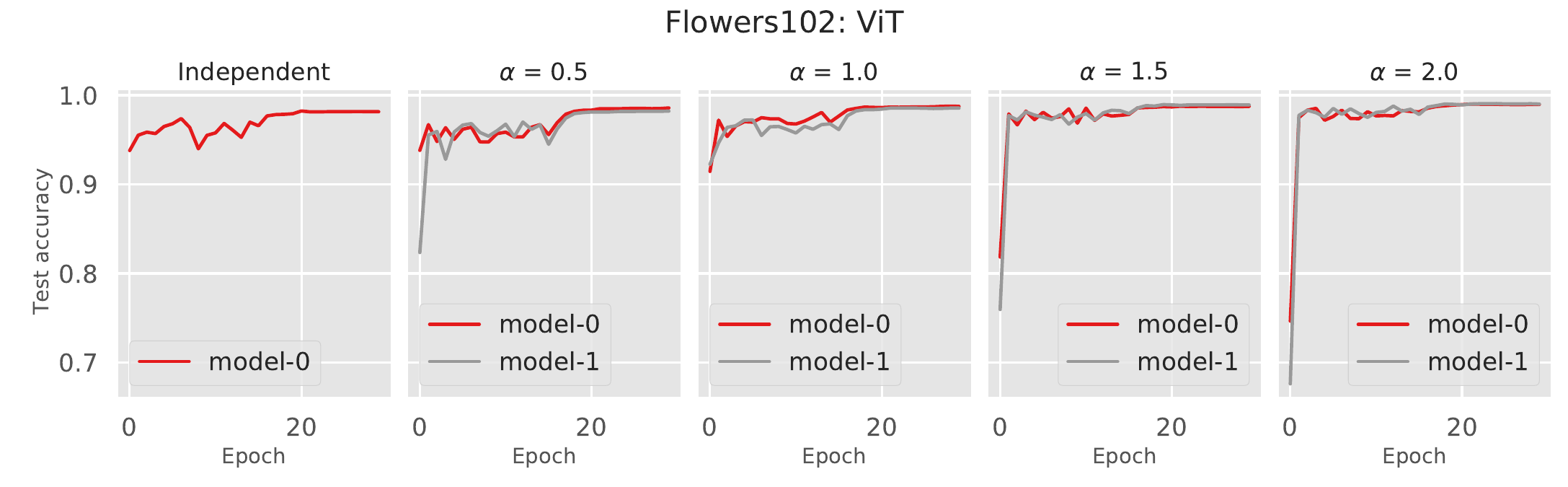}
}
\\
\subfloat{
\includegraphics[width=0.74\textwidth]{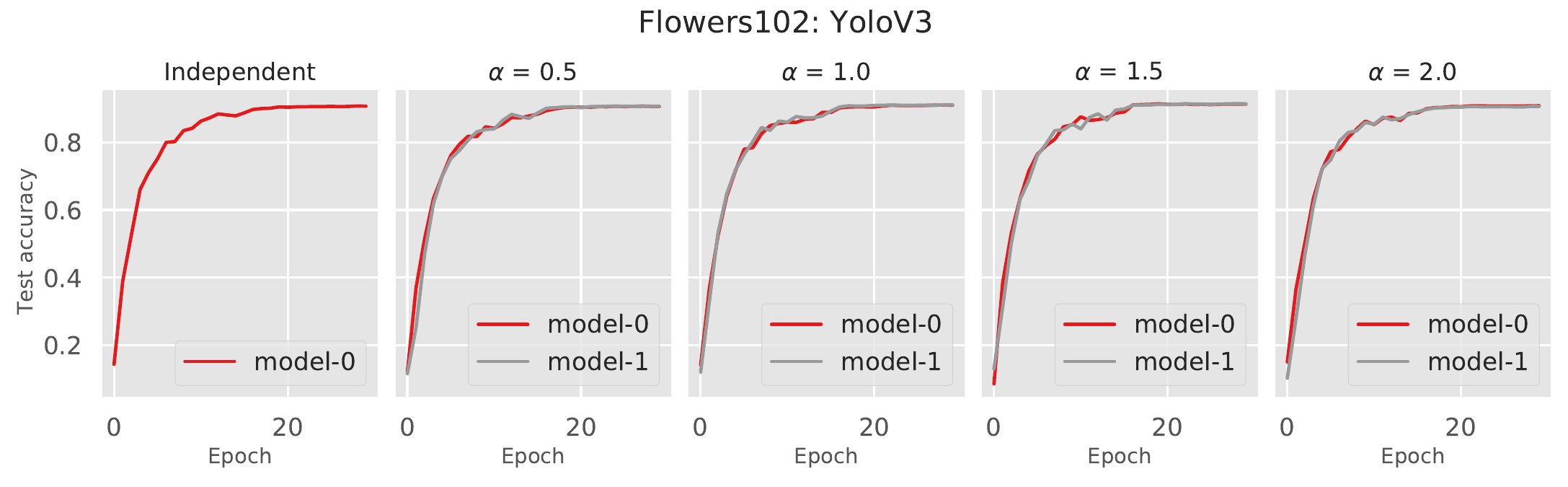}
}
\\
\subfloat{
\includegraphics[width=0.74\textwidth]{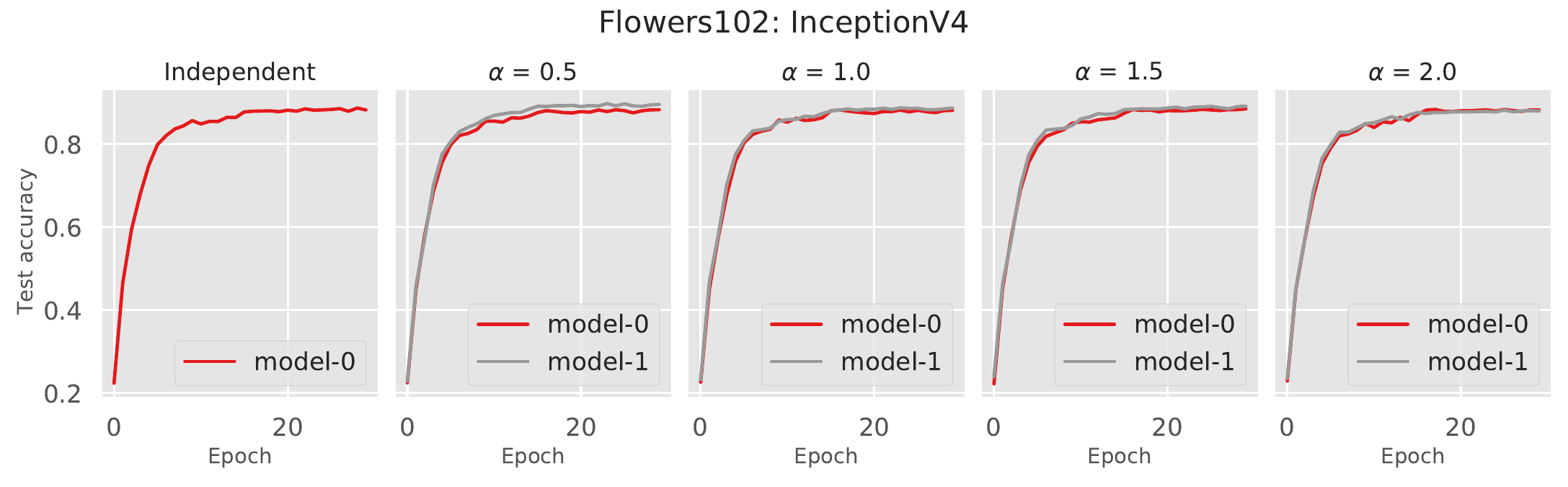}
}
\caption{Test accuracy of \rdml{} for various $\alpha$ on the \texttt{Flowers102} dataset.}
\label{fg:flowers102_val_acc}
\end{figure}

\cref{fg:flowers102_val_acc} exhibits the results of test accuracy on the \texttt{Flowers102} dataset with the three selected models.
For InceptionV4 and YoloV, the test accuracy of each model increases as the model is trained for more epochs.
Regarding ViT, we can observe that the test accuracy of the independent model tends to fluctuate in the first 15 epochs during training.
In contrast, we also notice that the accuracy curves become less fluctuating using \rdml{} with larger $\alpha$ values, which again shows the flexibility of tuning $\alpha$ in \rdml{} for better generalization.

%% file: relatedwork.tex

\section{Related Work}
\label{sec:relatedwork}
DML is a knowledge distillation scheme that transfers knowledge from from one or more deep neural networks to a designated one.
Unlike offline distillation, where knowledge is transferred from a pretrained model, DML allows multiple neural networks to collaborate and transfer knowledge during training, providing flexibility to train different or the same networks~\cite{gou2021knowledge}.
Due to its effectiveness, DML has been used in various contexts and applications, such as tracking visual objects~\cite{zhao2021deep}, machine translation~\cite{zhao2021mutual}, speech recognition~\cite{masumura2020end}, and COVID-19 recognition~\cite{zhang2022dcml}.
Recently, Yang et al.~\cite{yang2020mutualnet} proposed training a cohort of sub-(convolutional) networks with different configurations of network widths and input resolutions via DML to achieve accuracy-efficiency tradeoffs.
Park et al.~\cite{park2020diversified} applied DML to deep metric learning beyond classification tasks and demonstrated its effectiveness over individual models.
However, to the best of our knowledge, there is no theoretical analysis of convergence in DML available.

The \renyi{} divergence~\cite{Erven2014} has garnered increasing interest and has been utilized in a variety of applications.
For instance, it has been proposed to replace the KL divergence in variational inference~\cite{li2016renyi}. Examples show that the~\renyi{} divergence can control the regularization power of the inferred variational distribution.
Several related applications in cryptography~\cite{bai2018improved,Prest2017} also exist, as the~\renyi{} divergence deliver tighter security bounds.
In addition, it has been widely used as a powerful tool to analyze differential privacy~\cite{mironov2017renyi,wang2019subsampled} and has been utilized in the examination of human brain behaviors~\cite{sajid2022bayes}.
All of these illustrate the potential of this divergence in a broad range of tasks.

%% file: conclusion.tex

\section{Conclusion}
\label{sec:conclusion}
In this paper, we have revisited DML and proposed a revised paradigm, namely \rdml{}.
Our motivation stems from the enhanced flexibility offered by \renyi{} divergence, compared to the KL divergence used in DML.
The empirical results support our findings and demonstrate that \rdml{} has greater capacity than DML to improve model performance.
Moreover, we theoretically proved the convergence guarantee of the paradigm, showing that the learning procedure will converge with a bounded bias.
This might help the learned parameters escape from a narrow optimum to a wider one in practice, particularly in cases where models tend to overfit the training data.

In regard to future research, one could examine the generalization error bounds for~\rdml{} to theoretically explore why it can learn better generalized models.
Additionally, investigating online hyperparameter tuning for the controlling parameter $\alpha$ is another potential avenue to explore.

%% file: appendix.tex
\appendix

\section{Derivations}
We show the detailed derivation of~\cref{eq:dml_div} here.
Let us denote two sequences of distributions $\{\mathbb{P}_1, \dots, \infty\}$ and $\{\mathbb{Q}_1, \dots, \infty\}$.
The~\renyi{} divergence maintains the additivity~\cite{Erven2014} such that
\[
\renyidiv \left( \prod_n \mathbb{P}_n || \prod_n \mathbb{Q}_n \right) = \sum_n \renyidiv(\mathbb{P}_n || \mathbb{Q}_n)
\]
for $\alpha \in [0, 1) \cup (1, \infty)$ and $N \le \infty$.

Recall that $p(\bm{\mu} | x, \bm{\theta}) \propto p(\bm{\mu} | x, \bm{\theta}) p(x)$ where $p(x)$ is a constant.
We also define a set $\bm{M} = \{ \bm{\mu}_1, \dots, \bm{\mu}_n \}$.
Considering that $\renyidiv [p(\X) || p(\X)] = 0$, we show
\begin{align*}
\obj^{div}_{k}
&\coloneqq \frac{1}{K-1} \sum_{j \in \m{s}_{\neg k}} \renyidiv [\mathbb{P}(\bm{M} | \X, \bm{\theta}_{j}) || \mathbb{P}(\bm{M} | \X, \bm{\theta}_k)] \nonumber \\
&= \frac{1}{K-1} \sum_{j \in \m{s}_{\neg k}} \renyidiv [\mathbb{P}(\bm{M}, \X | \bm{\theta}_{j}) || \mathbb{P}(\bm{M}, \X | \bm{\theta}_k)] \\
&= \frac{1}{K-1} \sum_{j \in \m{s}_{\neg k}} \renyidiv [ \prod_n \mathbb{P}(\bm{\mu}_n, x_n | \bm{\theta}_{j}) || \prod_n \mathbb{P}(\bm{\mu}_n, x_n | \bm{\theta}_k) ] \nonumber \\
&= \frac{1}{K-1} \sum_{j \in \m{s}_{\neg k}} \sum_n \renyidiv [ \mathbb{P}(\bm{\mu}_n, x_n | \bm{\theta}_{j}) || \mathbb{P}(\bm{\mu}_n, x_n | \bm{\theta}_k) ] \nonumber \\
&= \frac{1}{K-1} \sum_{j \in \m{s}_{\neg k}} \sum_{n} \renyidiv [ \mathbb{P}(\bm{\mu}_n | x_n, \bm{\theta}_{j}) || \mathbb{P}(\bm{\mu}_n | x_n, \bm{\theta}_k)] \,.
\end{align*}

\section{Unbiased Gradient Estimator}
\begin{proof}[Proof of~\cref{prop:unbiased}]
In SGD, every student will first sample a data point $d$ and a peer $j$ to derive their gradients with regard to $\bm{\theta}_{k, t}$.
It then proceeds with
\begin{align*}
  \bm{\theta}_{k, t + 1} = \bm{\theta}_{k, t} - \eta_t \nabla \obj_k(d, j, \bm{\theta}_{k, t}) \,.
\end{align*}
The expectation of the gradients is unbiased, i.e.,
\begin{align*}
\E [\nabla \obj_k(d, j, \bm{\theta}_{k, t})]
&= \sum_{d \in \data} \sum_{j \in \m{s}_{\neg k}} \underbrace{p(d = (x, y))}_{\frac 1 N} \underbrace{p(j)}_{\frac 1 {K-1}} \nabla_{\bm{\theta}_{k, t}} \{ \obj^{base}(d, \bm{\theta}_{k, t}) + \renyidiv[\mathbb{P} (\bm{\mu} | x,  \bm{\theta}_{j, t}) || \mathbb{P}(\bm{\mu} | x_n, \bm{\theta}_{k, t})] \} \\
&= \sum_{d \in \data} p(d) \nabla_{\bm{\theta}_{k, t}} \left( \obj^{base}_{k}(d_n, \bm{\theta}_{k, t}) + \sum_{j \in \m{s}_{\neg k}} p(j) \renyidiv[\mathbb{P}(\bm{\mu} | x, \bm{\theta}_{j, t}) || \mathbb{P}(\bm{\mu} | x, \bm{\theta}_{k, t})] \right) \\
&= \nabla_{\bm{\theta}_{k, t}} \frac 1 N \left ( \sum_{d \in \data} \obj^{base}_{k}(d, \bm{\theta}_{k, t}) +  \frac {1} {K-1} \sum_{j \in \m{s}_{\neg k}} \renyidiv[\mathbb{P}(\bm{\mu} | x, \bm{\theta}_{j, t}) || \mathbb{P}(\bm{\mu} | x, \bm{\theta}_{k, t})] \right ) \\
&= \nabla \obj_k(\bm{\theta}_{k, t}) \,.
\end{align*}
We finish the proof for our claim.
\end{proof}

\section{Theoretical Analysis of the Convergence Guarantees}
In this analysis, we focus on the convergence guarantees of \rdml{} for nonconvex optimization tasks.
This analysis is divided into two parts:
\begin{enumerate}
\item
We analyze the continuity and smoothness of the loss functions;
\item
We prove the convergence guarantees of the RDML algorithm.
\end{enumerate}

\subsection{Lipschitz Continuity and Smoothness}
As our focus is the convergence guarantee for the RDML on  nonconvex problems, we have to first validate that the RDML objective follows the smoothness.
The function $g(\cdot)$ depends on the task but is assumed to be smooth; however, we have to check the smoothness for the divergence part in the RDML loss.
To this end, we first analyze the Lipschitz Continuity and smoothness of the~\renyi{} divergence.

\begin{lemma}
\label{prop:b-smooth-bound}
Let $\varphi = \min \left\{ \tau^{-1}, \left(\tau e^{(M-1) \tau}\right)^{-\alpha} M \right\} L$.
Suppose $\bm{\mu}$, $x$, and $\bm{\theta}_j$ are fixed.
Under~\cref{asmp:cont,asmp:smooth}, the~\renyi{} divergence related term
\[
| \renyidiv[\mathbb{P}(\bm{\mu} | x, \bm{\theta}_j) || \mathbb{P}(\bm{\mu} | x, \bm{\theta}_k)]
- \renyidiv[\mathbb{P}(\bm{\mu} | x, \bm{\theta}_j) || \mathbb{P}(\bm{\mu} | x, \bm{\theta}_k')] |
\le \varphi \| \bm{\theta}_k - \bm{\theta}_k' \| \,,
\]
i.e., $\renyidiv[\mathbb{P}(\bm{\mu} | x, \bm{\theta}_j) || \mathbb{P}(\bm{\mu} | x, \bm{\theta}_k)$ is $\varphi$-Lipschitz in $\bm{\theta}_k$.
\end{lemma}
\begin{proof}
We denote a normalizing constant by $Z$ where $Z = \int p(\mu | x, \bm{\theta}_j)^{\alpha} \mathbb{P}(\mu| x,  \bm{\theta}_k)^{1-\alpha} d \mu$.
For any $\bm{\mu}, x, \bm{\theta}_k$, we obtain
\begin{align*}
\nabla_{\bm{\theta}_k} \renyidiv[\mathbb{P}(\bm{\mu}| x, \bm{\theta}_j) || \mathbb{P}(\bm{\mu}| x, \bm{\theta}_k)]
&= \nabla_{\bm{\theta}_k} \frac 1 {\alpha-1} \log \int p(\mu | x, \bm{\theta}_j)^{\alpha} p(\mu| x,  \bm{\theta}_k)^{1-\alpha} d \mu \\
&= \frac {\nabla_{\bm{\theta}_k} \int p(\mu | x, \bm{\theta}_j)^{\alpha} p(\mu| x,  \bm{\theta}_k)^{1-\alpha} d \mu} {(\alpha-1) \int p(\mu | x, \bm{\theta}_j)^{\alpha} p(\mu| x,  \bm{\theta})^{1-\alpha} d \mu}  \\
&= \frac 1 {(\alpha-1) Z} \int p(\mu | x, \bm{\theta}_j)^{\alpha} \nabla_{\bm{\theta}_k} p(\mu| x,  \bm{\theta}_k)^{1-\alpha} d \mu \,,
\end{align*}
via the chain rule.
We get
\begin{align}
\label{eq:renyi_norm_ckpt}
\| \nabla_{\bm{\theta}_k} \renyidiv[\mathbb{P}_j(\bm{\mu}| x, \bm{\theta}_j) || \mathbb{P}_k(\bm{\mu}| x, \bm{\theta}_k)] \|
&= \left \| \frac 1 {(\alpha-1) Z} \int p(\mu | x, \bm{\theta}_j)^{\alpha} \nabla_{\bm{\theta}_k} p(\mu| x,  \bm{\theta}_k)^{1-\alpha} d \mu \right\| \nonumber \\
&\le \frac 1 {Z |\alpha-1|} \int \| p(\mu | x, \bm{\theta}_j)^{\alpha} \nabla_{\bm{\theta}_k} p(\mu| x,  \bm{\theta}_k)^{1-\alpha} \| d\mu \nonumber \\
&= \frac 1 {Z |\alpha-1|} \int \| (1-\alpha) p(\mu | x, \bm{\theta}_j)^{\alpha} p(\mu| x, \bm{\theta}_k)^{-\alpha} \nabla_{\bm{\theta}_k} p(\mu| x,  \bm{\theta}_k) \| d \mu \nonumber \\
&= \frac 1 {Z} \int p(\mu | x, \bm{\theta}_j)^{\alpha} p(\mu| x, \bm{\theta}_k)^{-\alpha} \| \nabla_{\bm{\theta}_k} p(\mu| x,  \bm{\theta}_k) \| d \mu \,.
\end{align}
Note that
\begin{align}
\label{eq:z_upper1}
Z
&= \int p(\mu| x, \bm{\theta}_k) [p(\mu | x, \bm{\theta}_j)^{\alpha} p(\mu| x, \bm{\theta}_k)^{-\alpha}] d \nonumber \\
&\ge \min_{\mu} p(\mu| x, \bm{\theta}_k) \int p(\mu | x, \bm{\theta}_j)^{\alpha} p(\mu| x, \bm{\theta}_k)^{-\alpha} d \mu \nonumber \\
&\ge \tau \int p(\mu | x, \bm{\theta}_j)^{\alpha} p(\mu| x, \bm{\theta}_k)^{-\alpha} d \mu \,.
\end{align}
Consider that $\forall \mu: \nabla_{\bm{\theta}_k} p(\mu| x, \bm{\theta}_k)$ being $L$-Lipschitz in $\bm{\theta}_k$ implies $\forall \mu: \| \nabla_{\bm{\theta}_k} p(\mu| x, \bm{\theta}_k) \| \le L$.
\cref{eq:renyi_norm_ckpt} can be expressed as
\begin{align}
\label{eq:renyi_grad_bound0}
\| \nabla_{\bm{\theta}_k} \renyidiv[\mathbb{P}(\bm{\mu}| x, \bm{\theta}_j) || \mathbb{P}(\bm{\mu}| x, \bm{\theta}_k)] \|
&\le
\frac{1}{Z} \max_{\mu} \| \nabla_{\bm{\theta}_k} p( \mu | x, \bm{\theta}_k) \|  \int p(\mu | x, \bm{\theta}_j)^{\alpha} p(\mu| x, \bm{\theta}_k)^{-\alpha} d \mu \nonumber \\
&=
\frac {\int p(\mu | x, \bm{\theta}_j)^{\alpha} p(\mu| x, \bm{\theta}_k)^{-\alpha} d \mu}{Z} \max_{\mu} \| \nabla_{\bm{\theta}_k} p( \mu | x, \bm{\theta}_k) \| \nonumber \\
&\le
\tau^{-1} L \,.
\end{align}

Now, let us consider the case which takes the length of $|\bm{\mu}| = M$ into account.
This implies that $\forall \mu: \tau < p(\mu | \cdot) < 1 - (M-1) \tau < 1$ holds.
We extend~\cref{eq:renyi_norm_ckpt} to
\begin{align}
\label{eq:renyi_grad_bound}
\| \nabla_{\bm{\theta}_k} \renyidiv[\mathbb{P} (\bm{\mu}| x, \bm{\theta}_j) \| \mathbb{P}(\bm{\mu}| x, \bm{\theta}_k)] \|
&\le
\frac 1 Z \max_{\mu} \| \nabla_{\bm{\theta}_k} p( \mu | x, \bm{\theta}_k) \|  \int p(\mu | x, \bm{\theta}_j)^{\alpha} p(\mu| x, \bm{\theta}_k)^{-\alpha} d \mu \nonumber \\
&\le
\left( \frac{1 - (M-1) \tau}{\tau} \right)^\alpha M L \nonumber \\
&\le
\left( \frac{e^{- (M-1) \tau}}{\tau} \right)^\alpha M L \nonumber \\
&=
\left(\tau e^{(M-1) \tau}\right)^{-\alpha} M L \,.
\end{align}

Finally, combining~\cref{eq:renyi_grad_bound0,eq:renyi_grad_bound}, we achieve $\| \nabla_{\bm{\theta}_k} \renyidiv[ \mathbb{P}_j(\bm{\mu}| x, \bm{\theta}_j) || \mathbb{P}_k(\bm{\mu}| x, \bm{\theta}_k)] \| \le \varphi$ which indicates that the derivatives of interest are bounded given the assumptions.
By the mean value theorem, we prove our claim.
\end{proof}

The following corollary is a natural extension of~\cref{prop:b-smooth-bound}.
\begin{corollary}
\label{co:bounded_div_grad}
Based on~\cref{asmp:cont,asmp:smooth}, the~\renyi{} divergence term $\frac 1 N \renyidiv[\mathbb{P}_j(\bm{\mu} | \m{x}, \bm{\theta}_j) || \mathbb{P}_k(\bm{\mu} | \m{x}, \bm{\theta}_k)]$ is $\varphi$-Lipschitz in $\bm{\theta}_k$ given $\bm{\mu}$, $\m{x}$, and $\bm{\theta}_j$ fixed.
\end{corollary}
\begin{proof}
We apply the results from~\cref{prop:b-smooth-bound} (precisely~\cref{eq:renyi_grad_bound}) and write
\begin{align}
\left \| \frac 1 N \nabla_{\bm{\theta}_k} \renyidiv[ \mathbb{P}_j(\bm{\mu} | \m{x}, \bm{\theta}_j) || \mathbb{P}_k(\bm{\mu} | \m{x}, \bm{\theta}_k)] \right \|
\le \frac 1 N \sum_{n} \left \|\nabla_{\bm{\theta}_k} \renyidiv [ \mathbb{P}_j(\bm{\mu} | x_n, \bm{\theta}_j) || \mathbb{P}_k(\bm{\mu} | x_n, \bm{\theta}_k)] \right \|
\le \varphi \,.
\end{align}
Here we finish the proof.
\end{proof}

To prove~\cref{thm:q_smooth}, we need the following intermediate result.
\begin{lemma}
\label{thm:div-loss-smooth}
Let $\hat{L} = |\alpha-1| \varphi^2 + \alpha L^2 + \tau H$.
Suppose~\cref{asmp:smooth,asmp:cont} hold.
Considering the general \renyi{} divergence, the divergence loss $\obj^{div}_k$ for every student $k$ is $\hat{L}$-smooth in its parameter $\bm{\theta}_k$ given $\{\bm{\theta}_j\}_{j \in \m{s}_{\neg k}}$ fixed.
\end{lemma}

\begin{proof}
Our strategy is to derive the second order derivative of $\bm{\theta}_k$ in $\renyidiv[\mathbb{P}_j(\bm{\mu} | \m{x}, \bm{\theta}_j) || \mathbb{P}_k(\bm{\mu} | \m{x}, \bm{\theta}_k)]$ and show it is bounded.
Hence, by the mean value theorem, we can prove our claim.

We now derive the corresponding seconder order derivative.
Applying the chain rule, we obtain
\begin{align*}
&\nabla^2_{\bm{\theta}_k} \renyidiv[\mathbb{P}_j(\bm{\mu} | \m{x}, \bm{\theta}_j) || \mathbb{P}_k(\bm{\mu} | \m{x}, \bm{\theta}_k)] \\
&= \nabla_{\bm{\theta}_k} \frac 1 {(\alpha-1) Z} \int p(\mu | x, \bm{\theta}_j)^{\alpha} \nabla_{\bm{\theta}_k} p(\mu| x,  \bm{\theta}_k)^{1-\alpha} d \mu \\
&= \frac{1}{\alpha-1} \left[-\frac 1 {Z^2}  \int p(\mu | x, \bm{\theta}_j)^{\alpha} \nabla_{\bm{\theta}_k} p(\mu| x, \bm{\theta}_k)^{1-\alpha} d \mu \left(\nabla_{\bm{\theta}_k} Z \right)^{\top} + \frac 1 Z \int p(\mu | x, \bm{\theta}_j)^{\alpha} \nabla_{\bm{\theta}_k}^2 p(\mu| x, \bm{\theta}_k)^{1-\alpha} d \mu \right] \\
&= \frac{1}{\alpha-1} \left[ - \tilde{\bm{\theta}} \tilde{\bm{\theta}}^\top + \frac 1 Z \int p(\mu | x, \bm{\theta}_j)^{\alpha} \nabla_{\bm{\theta}_k}^2 p(\mu| x, \bm{\theta}_k)^{1-\alpha} d \mu \right]
\end{align*}
where $\tilde{\bm{\theta}} = \frac{1}{Z} \int p(\mu | x, \bm{\theta}_j)^{\alpha} \nabla_{\bm{\theta}_k} p(\mu| x, \bm{\theta}_k)^{1-\alpha} d \mu$.

For the first term, it follows from~\cref{thm:div-loss-smooth} that,
\begin{align}
\label{eq:renyi_smooth_lhs}
\left \| - \frac{1}{\alpha-1} \tilde{\bm{\theta}} \tilde{\bm{\theta}}^\top \right \|
&\le \frac{1}{|\alpha-1|} \| \tilde{\bm{\theta}} \| \| \tilde{\bm{\theta}} \| \nonumber \\
&= \frac{1}{|\alpha-1|} \left \| (\alpha-1) \nabla_{\bm{\theta}_k} \renyidiv[\mathbb{P}_j(\bm{\mu} | \m{x}, \bm{\theta}_j) || \mathbb{P}_k(\bm{\mu} | \m{x}, \bm{\theta}_k)] \right \|^2 \nonumber \\
&\le |\alpha-1| \varphi^2 \,.
\end{align}
With respect to the second term, we first derive the following sub-term:
\begin{align}
\label{eq:second_order_derivative}
&\| \nabla_{\bm{\theta}_k}^2 p(\mu| x, \bm{\theta}_k)^{1-\alpha} \| \nonumber \\
&= \| \nabla_{\bm{\theta}_k} (1-\alpha) p(\mu| x, \bm{\theta}_k)^{-\alpha} \nabla_{\bm{\theta}_k} p(\mu| x, \bm{\theta}_k) \| \nonumber \\
&= |\alpha-1| \| -\alpha p(\mu| x, \bm{\theta}_k)^{-1-\alpha}\nabla_{\bm{\theta}_k}  p(\mu| x, \bm{\theta}_k) \nabla_{\bm{\theta}_k} p(\mu| x, \bm{\theta}_k)^\top +  p(\mu| x, \bm{\theta}_k)^{-\alpha} \nabla_{\bm{\theta}_k}^2p(\mu| x, \bm{\theta}_k) \| \nonumber \\
&\le |\alpha-1| p(\mu| x, \bm{\theta}_k)^{-\alpha} \| -\alpha p(\mu| x, \bm{\theta}_k)^{-1}\nabla_{\bm{\theta}_k} p(\mu| x, \bm{\theta}_k) \nabla_{\bm{\theta}_k} p(\mu| x, \bm{\theta}_k)^\top +  \nabla_{\bm{\theta}_k}^2p(\mu| x, \bm{\theta}_k) \| \nonumber \\
&\le |\alpha-1| p(\mu| x, \bm{\theta}_k)^{-\alpha} \left( \| -\alpha p(\mu| x, \bm{\theta}_k)^{-1} \| \| \nabla_{\bm{\theta}_k} p(\mu| x, \bm{\theta}_k) \|^2 + \| \nabla_{\bm{\theta}_k}^2 p(\mu| x, \bm{\theta}_k) \| \right) \nonumber \\
&\le |\alpha-1| p(\mu| x, \bm{\theta}_k)^{-\alpha} \left( \frac{\alpha}{\tau} L^2 + H \right) \,.
\end{align}
As~\cref{asmp:cont} satisfies, for any $\mu$ and $x$, $\| \nabla_{\bm{\theta}_k} p(\mu| x, \bm{\theta}_k) \| \le L$ and $\| \nabla_{\bm{\theta}_k}^2 p(\mu| x, \bm{\theta}_k) \| \le H$ hold, which validates the last line of the above inequality.
Following~\cref{eq:second_order_derivative}, we have
\begin{align}
\label{eq:renyi_smooth_rhs}
& \left \| \frac{1}{|1 - \alpha|} \frac 1 Z \int p(\mu | x, \bm{\theta}_j)^{\alpha} \nabla_{\bm{\theta}_k}^2 p(\mu| x, \bm{\theta}_k)^{1-\alpha} d \mu \right \| \nonumber \\
&\le \frac{1}{|1 - \alpha|} \frac 1 Z \int p(\mu | x, \bm{\theta}_j)^{\alpha} \left \| \nabla_{\bm{\theta}_k}^2 p(\mu| x, \bm{\theta}_k)^{1-\alpha} \right \| d \mu \nonumber \\
&\le \frac{1}{|1 - \alpha|} \frac 1 Z \int p(\mu | x, \bm{\theta}_j)^{\alpha} |\alpha-1| p(\mu| x, \bm{\theta}_k)^{-\alpha} \left( \frac{\alpha}{\tau} L^2 + H \right) d \mu \nonumber \\
&= \left( \frac{\alpha}{\tau} L^2 + H \right) \left[ \frac 1 Z \int p(\mu | x, \bm{\theta}_j)^{\alpha} p(\mu| x, \bm{\theta}_k)^{-\alpha} d \mu \right] \nonumber \\
&\le \alpha L^2 + \tau H \,,
\end{align}
where the last two lines are based on~\cref{eq:z_upper1}.


Now, we combine the two terms~\cref{eq:renyi_smooth_lhs,eq:renyi_smooth_rhs} and acquire that
\begin{align*}
\| \nabla^2_{\bm{\theta}_k} \renyidiv[\mathbb{P}_j(\bm{\mu} | \m{x}, \bm{\theta}_j) || \mathbb{P}_k(\bm{\mu} | \m{x}, \bm{\theta}_k)] \|
&\le |\alpha-1| \varphi^2 + \alpha L^2 + \tau H
= \hat{L} \,.
\end{align*}
Thus,
\begin{align*}
\| \nabla^2_{\bm{\theta}_k} \obj^{div}_k \|
&= \left \| \frac 1 {N(K-1)} \sum_n \sum_{j \in \m{s}_{\neg k}} \nabla^2_{\bm{\theta}_k} \renyidiv[\mathbb{P}_j(\bm{\mu} | \m{x}, \bm{\theta}_j) || \mathbb{P}_k(\bm{\mu} | \m{x}, \bm{\theta}_k)] \right \| \\
&\le \frac 1 {N(K-1)} \sum_n \sum_{j \in \m{s}_{\neg k}}\left \| \nabla^2_{\bm{\theta}_k} \renyidiv[\mathbb{P}_j(\bm{\mu} | \m{x}, \bm{\theta}_j) || \mathbb{P}_k(\bm{\mu} | \m{x}, \bm{\theta}_k)] \right \| \\
&\le \hat{L} \,.
\end{align*}
By the mean value theorem, one can assure the $\hat{L} $-smoothness for $\obj^{div}_k$ in $\bm{\theta}_k$.
\end{proof}
A special case for $\alpha \to 1$, which leads to the KL divergence, follows the corollary described next.
\begin{corollary}
Suppose~\cref{asmp:smooth,asmp:cont} hold.
When considering DML, the divergence loss $\obj^{div}_k$ for every student $k$ is $\left( L^2   + {\tau} H \right)$-smooth in its parameter $\bm{\theta}_k$ where $\bm{\theta}_j$  is fixed for every $j \in \m{s}_{\neg k}$ .
\end{corollary}
\begin{proof}
One may calculate $\lim_{\alpha \to 1} \hat{L}$ to achieve the desired result.
We omit the proof here.
\end{proof}

Finally, we are able to prove~\cref{thm:q_smooth}.
\begin{proof}[Proof of~\cref{thm:q_smooth}]
This is a straightforward application of~\cref{thm:div-loss-smooth}.
Recall that~\cref{asmp:cont,asmp:smooth} satisfy.
We write
\begin{align*}
&\left \| \nabla_{\bm{\theta}_k} \obj_k - \nabla_{\bm{\theta}_k'} \obj_k \right \| \\
&= \left \| \frac 1 N \left\{ \sum_{n} \nabla_{\bm{\theta}_{k}} \obj^{base}_{k}(d_n, \bm{\theta}_{k}) - \nabla_{\bm{\theta}_{k}'} \obj^{base}_{k}(d_n, \bm{\theta}_{k}') \right. \right. \\
&\left. \left. \quad + \frac 1 {K-1} \sum_{j \in \m{s}_{\neg k}} \nabla_{\bm{\theta}_{k}} \renyidiv[\mathbb{P}_j(\bm{\mu} | x_n,  \bm{\theta}_{j}) || \mathbb{P}_k(\bm{\mu} | x_n, \bm{\theta}_k)] - \nabla_{\bm{\theta}_j} \renyidiv[ \mathbb{P}_j(\bm{\mu} | x_n, \bm{\theta}_{j}) || \mathbb{P}_k(\bm{\mu} | x_n, \bm{\theta}_k')] \right\} \right \| \\
&\le \frac 1 N \sum_{n} \| \nabla_{\bm{\theta}_{k}} \obj^{base}_{k}(d_n, \bm{\theta}_{k}) - \nabla_{\bm{\theta}_{k}'} \obj^{base}_{k}(d_n, \bm{\theta}_{k}') \| \\
&\quad + \frac 1 {N(K-1)} \sum_{j \in \m{s}_{\neg k}} \left \| \nabla_{\bm{\theta}_k} \renyidiv[ \mathbb{P}_j(\bm{\mu} | x_n,  \bm{\theta}_{j}) || \mathbb{P}_k(\bm{\mu} | x_n, \bm{\theta}_k)] - \nabla_{\bm{\theta}_k'} \renyidiv[ \mathbb{P}_j(\bm{\mu} | x_n,  \bm{\theta}_{j}) || \mathbb{P}_k(\bm{\mu} | x_n, \bm{\theta}_k')] \right \| \\
&\le \frac 1 N \left ( V \|\bm{\theta}_{k} - \bm{\theta}_{k}'\| + \frac 1 {K-1} \sum_{j \in \m{s}_{\neg k}} \hat{L} \|\bm{\theta}_{k} - \bm{\theta}_{k}'\| \right) \\
&= \left( V + \hat{L} \right)  \|\bm{\theta}_{k} - \bm{\theta}_{k}'\| \,.
\end{align*}
Setting $W = V + \hat{L} = V + |\alpha-1| \varphi^2 + {\alpha} L^2 + \tau H$ gives the desired result.
\end{proof}

\subsubsection{Proof of~\cref{lem:bounded_noise}}
We now proceed to prove~\cref{lem:bounded_noise}.
\begin{proof}[Proof of~\cref{lem:bounded_noise}]
Implied by~\cref{eq:renyi_grad_bound}, we have
\begin{align}
\| \nabla \obj_k^{div} (d, j, \bm{\theta}_k) \| \le \psi
\implies
\| \nabla \obj_k^{div} (d, j, \bm{\theta}_k) \|^2 \le \psi^2
\implies
\E \left[ \| \nabla \obj_k^{div}(d, j, \bm{\theta}_k) \|^2 \right] \le \psi^2 \,.
\end{align}
In combination with~\cref{asmp:bounded_grad}, we get
\begin{align*}
\E \left[\| \nabla \obj_k(d, j, \bm{\theta}_k) \|^2 \right]
& =
\E \left[\| \nabla \obj_k^{base}(d, \bm{\theta}_k) + \nabla \obj_k^{div} (d, j, \bm{\theta}_k) \|^2 \right] \\
&\le
\E \left[2\| \nabla \obj_k^{base}(d, \bm{\theta}_k) \|^2 + 2\|\nabla \obj_k^{div} (d, j, \bm{\theta}_k) \|^2 \right] \\
&=
2\tilde{\sigma}^2 + 2\E \left[\|\nabla \obj_k^{div} (d, j, \bm{\theta}_k) \|^2 \right] \\
&\le
2\tilde{\sigma}^2 + 2\varphi^2 \\
&\eqqcolon
\sigma^2 \,.
\end{align*}
This result applies to any time $t$.
We finish the proof.
\end{proof}

\subsection{Convergence of the Algorithm}
Here, we decompose the steps of the algorithm for a detailed analysis on the convergence rate.
At a specific loop, each student optimizes toward an independent objective $\obj_{k}$.
Although no extra steps are imposed, bias is generated during the process of learning from each other.
Therefore, we decompose each loop into two sub-loops with regard to the objective change.
The steps can be viewed as two sub-loops:
\begin{enumerate}
\item the active change in the loss, $\obj_k^{base}$, due to gradient descent;
\item the passive change in $\obj_k^{div}$ due to peers' change in their parameters, i.e., the~\renyi{} divergence between every pair of students changes when every parameter $\bm{\theta}_{k}$ changes.
\end{enumerate}
For a timestamp $t$, we write $\obj_{k, t}$ the entire loss of student $k$ with regard to the first sub-loop and $\tilde{\obj}_{k, t}$ for the second sub-loop which will be the starting point for $t+1$, i.e., $\tilde{\obj}_{k, t} \equiv \obj_{k, t+1}$.

In the remaining text, we use the notations $\nabla \obj_k(d, j, \bm{\theta}_{k, t})$ and $\nabla_{\bm{\theta}_{k, t}} \obj_k(d, j, \bm{\theta}_{k, t})$ interchangeably.
Given the $W$-smoothness of a single loss function for a certain $k$, we have
\begin{align*}
\tilde{\obj}_k(\bm{\theta}_{k, t + 1})
&\le \obj_k(\bm{\theta}_{k, t}) + [\tilde{\obj}_k(d, j, \bm{\theta}_{k, t + 1}) - \obj_k(d, j, \bm{\theta}_{k, t})]^\top \nabla \obj_k (\bm{\theta}_{k, t})+ \eta_{t}^2 \frac {W} 2 \| \nabla \obj_k(d, j, \bm{\theta}_{k, t}) \|^2 \\
&= \obj_k(\bm{\theta}_{k, t}) - \eta_t \nabla \obj_k(d, j, \bm{\theta}_{k, t})^\top \nabla \obj_k (\bm{\theta}_{k, t}) + \eta_{t}^2 \frac {W} 2 \| \nabla \obj_k(d, j, \bm{\theta}_{k, t}) \|^2 \,.
\end{align*}
which is standard result~\cite{nesterov2003introductory}.
Taking the expectation over $d$ and $j$, we get
\begin{align*}
\E [\tilde{\obj}_k(\bm{\theta}_{k, t + 1})]
&\le \obj_k(\bm{\theta}_{k, t}) - \eta_t \E[\nabla \obj_k (d, j, \bm{\theta}_{k, t}) ]^\top \obj_k (\bm{\theta}_{k, t}) + \eta_{t}^2 \frac {W} 2 \E \| \nabla \obj_k(d, j, \bm{\theta}_{k, t}) \|^2  \\
&= \obj_k(\bm{\theta}_{k, t}) - \eta_t \| \nabla \obj_k (\bm{\theta}_{k, t})\|^2 + \eta_{t}^2\frac {W} 2 \E \| \nabla \obj_k(d, j, \bm{\theta}_{k, t}) \|^2
\end{align*}
where $\E [\tilde{\obj}_k(\bm{\theta}_{k, t + 1})]$ denotes the expected value of $\tilde{\obj}_k(\bm{\theta}_{k, t + 1})$ conditional on the previous iteration $t$.
Hence, we obtain
\begin{align}
\label{eq:dml-ineq-before}
\eta_t \| \nabla \obj_k (\bm{\theta}_{k, t})\|^2
&\le \obj_k(\bm{\theta}_{k, t}) - \E [\tilde{\obj}_k(\bm{\theta}_{k, t + 1})] + \eta_t^2 \frac {W} 2 \E \| \nabla \obj_k(d, j, \bm{\theta}_{k, t}) \|^2 \,.
\end{align}
However, after each student update their $\bm{\theta}$, the~\renyi{} divergence term may also change for each student, and achieve $\obj_k(\bm{\theta}_{k, t + 1})$ rather than $\tilde{\obj}_k(\bm{\theta}_{k, t + 1})$ where
\begin{align}
\obj_k(\bm{\theta}_{k, t+1})
&= \obj_k^{base}(\bm{\theta}_{k, t + 1}) + \frac 1 {K-1} \sum_{j \in \m{s}_{\neg k}} \renyidiv [ \mathbb{P}_j(\bm{\mu}| x, {\color{red} \bm{\theta}_{j, t + 1} } ) || \mathbb{P}_k(\bm{\mu}| x,  \bm{\theta}_{k, t + 1})] \, ;
\end{align}
in contrast,
\begin{align}
\tilde{\obj}_k(\bm{\theta}_{k, t + 1})
&=\obj_k^{base}(\bm{\theta}_{k, t + 1}) + \frac 1 {K-1} \sum_{j \in \m{s}_{\neg k}} \renyidiv [ \mathbb{P}_j(\bm{\mu}| x, {\color{red} \bm{\theta}_{j, t} } ) || \mathbb{P}_k(\bm{\mu}| x, \bm{\theta}_{k, t + 1})] \, .
\end{align}
The red parts above have highlighted the difference between $\obj_k(\bm{\theta}_{k, t+1})$ and $\tilde{\obj}_k(\bm{\theta}_{k, t + 1})$.

Henceforth, we expand the following term
\begin{align*}
\obj_k(\bm{\theta}_{k, t}) - \E [\tilde{\obj}_k(\bm{\theta}_{k, t + 1})]
&= \obj_k^{base} (\bm{\theta}_{k, t}) + \frac 1 {N(K-1)} \sum_n \sum_{j \in \m{s}_{\neg k}} \renyidiv [ \mathbb{P}_j(\bm{\mu}| x_n, \bm{\theta}_{j, t} ) || \mathbb{P}_k(\bm{\mu}| x_n, \bm{\theta}_{k, t})] \\
&\quad - \E [\obj_k^{base} (\bm{\theta}_{k, t+1})] - \E [ \renyidiv \{ \mathbb{P}_j(\bm{\mu}| x, \bm{\theta}_{j, t}) || \mathbb{P}_k(\bm{\mu}| x, \bm{\theta}_{k, t + 1}) \}] \\
&= \obj_k^{base} (\bm{\theta}_{k, t}) - \E [\obj_k^{base} (\bm{\theta}_{k, t+1})] + \E[ \Psi_k(d, j, t) ]
\end{align*}
by defining
\begin{align*}
&\E[ \Psi_k(d, j, t) ] \\
&= \frac 1 {N(K-1)} \sum_n \sum_{j \in \m{s}_{\neg k}} \renyidiv [ \mathbb{P}_j(\bm{\mu}| x_n, \bm{\theta}_{j, t}) || \mathbb{P}_k(\bm{\mu}| x_n, \bm{\theta}_{k, t})] - \E [ \renyidiv [ \mathbb{P}_j(\bm{\mu}| x, \bm{\theta}_{j, t}) || \mathbb{P}_k(\bm{\mu}| x, \bm{\theta}_{k, t+1})]] \\
&= \frac 1 {N(K-1)} \sum_n \sum_{j \in \m{s}_{\neg k}} \left\{ \renyidiv [ \mathbb{P}_j(\bm{\mu}| x_n, \bm{\theta}_{j, t}) || \mathbb{P}_k(\bm{\mu}| x_n, \bm{\theta}_{k, t})] - \renyidiv [ \mathbb{P}_j(\bm{\mu}| x_n, \bm{\theta}_{j, t}) || \mathbb{P}_k(\bm{\mu}| x_n, \bm{\theta}_{k, t+1})] \right \} \,.
\end{align*}

Now,~\cref{eq:dml-ineq-before} can be equivalently expressed as
\begin{align}
\label{eq:grad_upperbound}
\eta_t \| \nabla \obj_k(\bm{\theta}_{k, t}) \|^2
&\le \obj_k^{base}(\bm{\theta}_{k, t}) - \E [\obj_k^{base}(\bm{\theta}_{k, t + 1})] + \E[ \Psi_k(d, j, t) ] + \eta_t^2 \frac {W} 2 \E \| \nabla \obj_k(d, j, \bm{\theta}_{k, t}) \|^2 \nonumber \\
&\le \obj_k^{base}(\bm{\theta}_{k, t}) - \E [\obj_k^{base}(\bm{\theta}_{k, t + 1})] + \E[ \Psi_k(d, j, t) ] + \eta_t^2 \frac {W} 2 \sigma^2 \,.
\end{align}
After $T$ loops with the progression through the law of iterated expectation, by telescoping sum, we arrive at
\begin{align*}
\sum_{t=1}^{T} \eta_t \E \| \nabla  \obj_k (\bm{\theta}_{k, t})\|^2
&\le \obj_k^{base}(\bm{\theta}_{k, 1}) - \E[ \obj_k^{base}(\bm{\theta}_{k, t + 1})] + \sum_{t=1}^T  \E[\Psi_k(d, j, t)] + \sum_{t=1}^T \eta_t^2 \frac {W\sigma^2} 2 \, .
\end{align*}
Recall that $\obj^{base}_{k ,*}$ is the global minimum of the loss function for the student, and thus $\obj^{base}_{k ,*} \le \E[ \obj_k^{base}(\bm{\theta}_{k, t + 1}) ]$ at any time $t$.
Also, given the fact that the min function is concave,
\begin{align}
\label{eq:dml-ineq-after}
\E \left[ \min_t \| \obj_k (\bm{\theta}_{k, t})\|^2 \right]
&\le \min_t \E \| \obj_k (\bm{\theta}_{k, t})\|^2 \nonumber \\
&\le \frac {\obj^{base}_k(\bm{\theta}_{k, 1}) - \obj^{base}_{k ,*}}{\sum_{t=1}^{T} \eta_t} + \frac{\sum_{t=1}^{T} \eta_t^2 W \sigma^2} {2\sum_{t=1}^{T} \eta_t} + \frac{\sum_{t=1}^{T} \E[\Psi_k(d, j, t)]} {\sum_{t=1}^{T} \eta_t} \,.
\end{align}
To analyze the convergence of the gradients, we have to focus on the above three therms.
The bias term (on the RHS) is apparently the most nontrivial term, while the other two are commonly seen in the conventional SGD analysis.
In the sequel, we analyze the upper bound of the bias terms and finally demonstrate its convergence guarantees.

\subsubsection{Analyzing the Bias Term}
To analyze the bias term, we require the following intermediate result.

\begin{lemma}
\label{thm:expected_bias}
Let us consider $\bm{\theta}_{k, t+1} = \bm{\theta}_{k, t} - \eta_t \nabla \obj_k(d, j, \bm{\theta}_{k, t})$ given any data point $d \in \data$.
Considering a data point $d$, for any student $k$ and $j$ where $k \ne j$,
\begin{align*}
\E[\Psi_k(d, j, t)]
\le \eta_t \varphi \E \| \nabla \obj_k (d, j, \bm{\theta}_{k, t}) \| \,.
\end{align*}
\end{lemma}
\begin{proof}
Employing the result in~\cref{prop:b-smooth-bound}, it is obvious that
\begin{align*}
\left | \renyidiv [\mathbb{P}_j(\bm{\mu}| x, \bm{\theta}_{j, t}) || \mathbb{P}_k(\bm{\mu}| x, \bm{\theta}_{k, t})] - \renyidiv [\mathbb{P}_j(\bm{\mu}| x, \bm{\theta}_{j, t}) || \mathbb{P}_k(\bm{\mu}| x, \bm{\theta}_{k, t+1})] \right |
&\le \varphi \| \bm{\theta}_{k, t} - \bm{\theta}_{k, t+1} \| \\
&= \eta_t \varphi \| \nabla \obj_k(d, j, \bm{\theta}_{k, t}) \| \,.
\end{align*}
given a fixed $k$ and a random data point $d$ and peer $j$.
Now, we show that
\begin{align}
\label{eq:expected_delta}
\E[\Psi_k(d, j, t)]
&= \E \{ \renyidiv [\mathbb{P}_j(\bm{\mu}| x, \bm{\theta}_{j, t}) || \mathbb{P}_k(\bm{\mu}| x, \bm{\theta}_{k, t})] - \renyidiv [\mathbb{P}_j(\bm{\mu}| x, \bm{\theta}_{j, t}) || \mathbb{P}_k(\bm{\mu}| x, \bm{\theta}_{k, t+1})] \} \nonumber \\
&\le \E \{ | \renyidiv [\mathbb{P}_j(\bm{\mu}| x, \bm{\theta}_{j, t}) || \mathbb{P}_k(\bm{\mu}| x, \bm{\theta}_{k, t})] - \renyidiv [\mathbb{P}_j(\bm{\mu}| x, \bm{\theta}_{j, t}) || \mathbb{P}_k(\bm{\mu}| x, \bm{\theta}_{k, t+1})] | \} \nonumber \\
&\le \eta_t \varphi \E \| \nabla \obj_k(d, j, \bm{\theta}_{k, t}) \| \,.
\end{align}
\end{proof}
The above result is always true in the worst case scenario.
However, it is possible to lower the bound for the average case.
We leave the discussion of the average case in~\cref{sec:avg-case}.

\subsubsection{Worst Case Convergence}
Finally, we are able to  prove our main theorem for convergence.
\begin{proof}[Proof of~\cref{thm:converge}]
Let us write $\Gamma_{k, t} = \obj_k(\bm{\theta}_t) - \E[\obj_k(\bm{\theta}_{t+1})] + \eta_t^2 \frac{W}{2} \sigma^2$, which extends from~\cref{eq:dml-ineq-after}.
We employ the telescoping sum on it over timestamps $t=1,\dots, T$ with the law of iterated expectation, and achieve $\sum_t \Gamma_{k, t} = \nabla \obj^{base}_{k ,*} + \sum_t \eta_t^2 \frac{W}{2} \sigma^2$.
Recall that $\Delta \obj^{base}_{k ,*} = \obj^{base}_k(\bm{\theta}_{k, 1}) - \obj^{base}_{k ,*}$; thus,
\[
\forall t: \Delta \obj^{base}_{k ,*} \ge \obj^{base}_k(\bm{\theta}_{k, 1}) - \E [\obj^{base}_k(\bm{\theta}_{k, t})] \,.
\]
By observing that $\eta_t \le \frac{1}{W \sqrt{T}}$,
\begin{align*}
\frac{ \Delta \obj^{base}_{k ,*}}{\sum_t \eta_t} + \frac{\sum_t \frac W 2 \eta_t^2 \sigma^2}{\sum_t \eta_t}
= \mathcal{O} \left( \frac{\Psi \obj^{base}_{k ,*}}{\eta \sqrt{T}} + \frac{\sigma^2}{2\sqrt{T}} \right)
= \mathcal{O} \left( \frac{1}{\sqrt{T}} \right) \,.
\end{align*}


On the other hand, one may find that
\begin{align*}
\E [\| \nabla \obj(\bm{\theta}_{k, t}) \|]
&= \sqrt{\E [\|\nabla \obj(\bm{\theta}_{k, t}) \|^2] - \mathbb{V}[\| \nabla \obj(\bm{\theta}_{k, t}) \|]} \\
&\le \sqrt{\sigma^2} \\
&= \sigma \\
&= \sqrt{2} \sqrt{\tilde{\sigma}^2 + \varphi^2} \\
&\le 2\max(\tilde{\sigma}, \varphi) \,,
\end{align*}
where $\mathbb{V}$ is the variance operator and it is always nonnegative.
Applying~\cref{eq:grad_upperbound,thm:expected_bias} directly, we can obtain
\begin{align}
\label{eq:upper2}
\E \left [ \min_t \| \nabla \obj(\bm{\theta}_{k, t}) \|^2 \right ]
&\le \frac{2[\Delta \obj^{base}_{k ,*} + \sum_t \eta_t^2 \frac{W}{2} \sigma^2]}{\sum_{t} \eta_t} + \frac{\sum_t \eta_t \varphi \sigma }{\sum_t \eta_t} \nonumber \\
&= \mathcal{O}\left( \frac 1 {\sqrt T} \right) + \mathcal{O}(2 \max(\varphi \tilde{\sigma}, \varphi^2)) \nonumber \\
&= \mathcal{O}\left( \frac 1 {\sqrt T} + 1 \right)\,.
\end{align}
The proof finishes here.
\end{proof}

\subsubsection{Average Case Convergence}
\label{sec:avg-case}
The key component for analyzing the average case is $\E [\Psi_k(d, j, t)]$.
We have the following theorem.
\begin{theorem}
Suppose~\cref{asmp:cont,asmp:smooth,asmp:bounded_grad} hold.
If there exists a number $T$ that allows $\sum_{t=1}^T \E [\Psi_k(d, j, t)] \le 0$, by picking $\eta_t \le \frac{1}{W \sqrt{T}}$, the expected gradient norm for student $k$ converges in $\mathcal{O}\left(\frac{1}{\sqrt{T}} \right)$.
\end{theorem}
\begin{proof}
One may apply the same techniques from the proof for~\cref{thm:converge}.
The detail is omitted.
\end{proof}

However, we cannot guarantee that $\E[\Psi_k(d, j, t)] \le 0$.
Now, we expand it from a perspective different from~\cref{thm:expected_bias}.
We first denote $\delta_{x, \mu, k, t} = p(\mu| x, \bm{\theta}_{k, t}) - p(\mu| x, \bm{\theta}_{k, t+1})$.

Considering a certain student $k$ with her peer $j$, we define
\begin{align*}
\phi_{x, \mu, k, t}
&= \frac{p(\mu| x, \bm{\theta}_{k, t})}{p(\mu| x, \bm{\theta}_{k, t+1})} \\
&= 1 + \frac{p(\mu| x, \bm{\theta}_{k, t}) - p(\mu| x, \bm{\theta}_{k, t+1})}{p(\mu| x, \bm{\theta}_{k, t+1})} \\
&= 1 + \frac{\delta_{x, \mu, k, t}}{p(\mu| x, \bm{\theta}_{k, t+1})} \,.
\end{align*}
Clearly, $\phi_{x, \mu, k, t}$ is a scale factor of each $p(\mu | \cdot)$ and thus is within $(0, \infty)$.
Let us recall
\begin{align*}
\Psi_k(d, j, t)
&= \renyidiv [ \mathbb{P}_j(\bm{\mu}| x, \bm{\theta}_{j, t}) || \mathbb{P}_k(\bm{\mu}| x, \bm{\theta}_{k, t})] - \renyidiv [\mathbb{P}_j(\bm{\mu}| x, \bm{\theta}_{j, t}) || \mathbb{P}_k(\bm{\mu}| x, \bm{\theta}_{k, t+1})] \,.
\end{align*}
There exists a number $\tilde{\phi}_{x, k, t}$, where $\min_{\mu} \phi_{x, \mu, k, t} \le \tilde{\phi}_{x, k, t} \le \max_{\mu} \phi_{x, \mu, k, t}$, satisfying:
\begin{align}
\Psi_k(d, j, t)
&=
\frac 1 {\alpha - 1} \log \frac{\int p(\mu | x, \bm{\theta}_{j, t})^\alpha p(\mu| x, \bm{\theta}_{k, t})^{1-\alpha} d\mu} {\int p(\mu | x, \bm{\theta}_{j, t})^\alpha p(\mu| x, \bm{\theta}_{k, t+1})^{1-\alpha} d\mu} \nonumber \\
&=
\frac 1 {\alpha - 1} \log \frac{\int p(\mu | x, \bm{\theta}_{j, t})^\alpha p(\mu| x, \bm{\theta}_{k, t+1})^{1-\alpha} \phi_{x, \mu, k, t}^{1-\alpha} d\mu} {\int p(\mu | x, \bm{\theta}_{j, t})^\alpha p(\mu| x, \bm{\theta}_{k, t+1})^{1-\alpha} d\mu} \nonumber \\
&=
\frac 1 {\alpha - 1} \log \frac{\tilde{\phi}_{x, k, t}^{1-\alpha} \int p(\mu | x, \bm{\theta}_{j, t})^\alpha p(\mu| x, \bm{\theta}_{k, t+1})^{1-\alpha} d\mu} {\int p(\mu | x, \bm{\theta}_{j, t})^\alpha p(\mu| x, \bm{\theta}_{k, t+1})^{1-\alpha} d\mu} \nonumber \\
&=
\frac 1 {\alpha-1} \log \tilde{\phi}_{x, k, t}^{1-\alpha} \\
&= - \log \tilde{\phi}_{x, k, t} \, .
\end{align}
To verify that $\min_{\mu} \phi_{x, \mu, k, t} \le \tilde{\phi}_{x, k, t} \le \max_{\mu} \phi_{x, \mu, k, t}$, one may prove it by contradiction, e.g., we assume $\tilde{\phi}_{x, k, t} \le \min_{\mu} \phi_{x, \mu, k, t}$ and $\tilde{\phi}_{x, k, t} \ge \max_{\mu} \phi_{x, \mu, k, t}$ in turn.
In addition, by knowing that $\sum_{\mu}\delta_{x, \mu, k, t} = 0$ implies $\min_{\mu} \phi_{x, \mu, k, t} \le 1$ and $\max_{\mu} \phi_{x, \mu, k, t} \ge 1$ both hold all the time, we get
\begin{align}
\label{eq:log_min_delta}
-\log \min_{\mu}{\phi}_{x, \mu, k, t}
&= - \log \left(\min_{\mu} 1 + \frac{\delta_{x, \mu, k, t}}{p(\mu | x, \bm{\theta}_{k, t+1})} \right) \ge 0 \,,
\end{align}
where the equality holds iff $p(\bm{\mu} | x, \bm{\theta}_{k, t+1}) \equiv p(\bm{\mu} | x, \bm{\theta}_{k, t})$.
Thus, the random variable $-\log \tilde{\phi}_{x, k, t}$ has the upper bound:
\begin{align*}
-\log \tilde{\phi}_{x, k, t} \le -\log \min_{\mu}{\phi}_{x, \mu, k, t}  \,.
\end{align*}
At a certain time $t$, we have
\begin{align*}
\E[\Psi_k(d, j, t)]
= \E \left[-\log \tilde{\phi}_{x, k, t} \right]
\le -\log \min_{\mu}{\phi}_{x, \mu, k, t}
\end{align*}
by the expectation of bounded random variable.
Since we know little about the density of $\log \tilde{\phi}_{x, k, t}$, we cannot ensure $\E[- \log \tilde{\phi}_{x, k, t}] \le 0$, as it can be either positive or negative.

Taking one step further, it is even harder to decide if $\E_t [\Psi_k(d, j, t)] \le 0$.
However, if the required information is given to ensure that $\sum_t \E [\Psi_k(d, j, t)] \le 0$ for a specific student $k$, the algorithm for this student $k$ converges in an unbiased fashion.

\section{Resources of Model and Software}
We list the public sites for the models and softwares that are used in our empirical study as follows.
\begin{itemize}
\item
\textbf{torchvision}: \url{https://pytorch.org/vision/stable/datasets.html}
\item
\textbf{torchtext}: \url{https://pytorch.org/text/stable/datasets.html}
\item
GoogleNet, ResNet34, VGG16: \url{https://github.com/weiaicunzai/pytorch-cifar100}
\item
\textbf{timm}: \url{https://rwightman.github.io/pytorch-image-models}
\item
fastText, CharCNN: \url{https://github.com/AnubhavGupta3377/Text-Classification-Models-Pytorch}
\end{itemize}

\section{Additional Generalization Results}
Since the generalization results for \texttt{CIFAR100} and \texttt{Flowers102} have been presented in the main text, we display the results for \texttt{CIFAR10} and \texttt{DTD} here.

ViT for \texttt{DTD} is a clear case that \rdml{} greatly improve the generalization performance over the original ViT model, for this task.
\begin{figure*}[h]
\centering
\subfloat{
\includegraphics[width=0.75\textwidth]{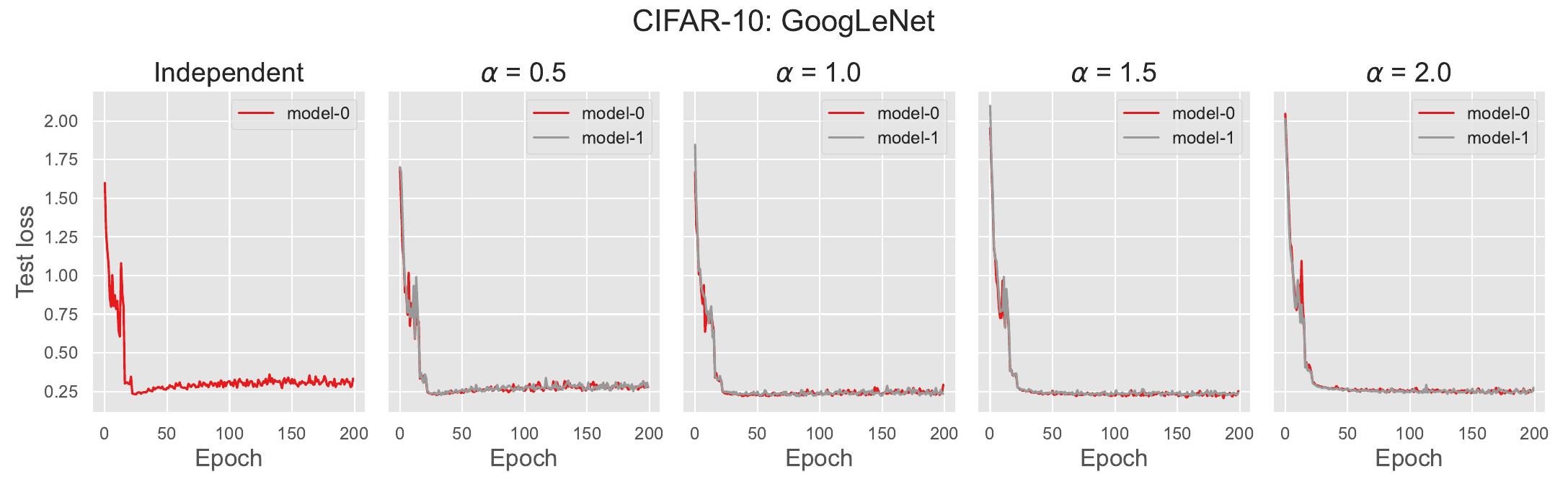}
}
\\
\subfloat{
\includegraphics[width=0.75\textwidth]{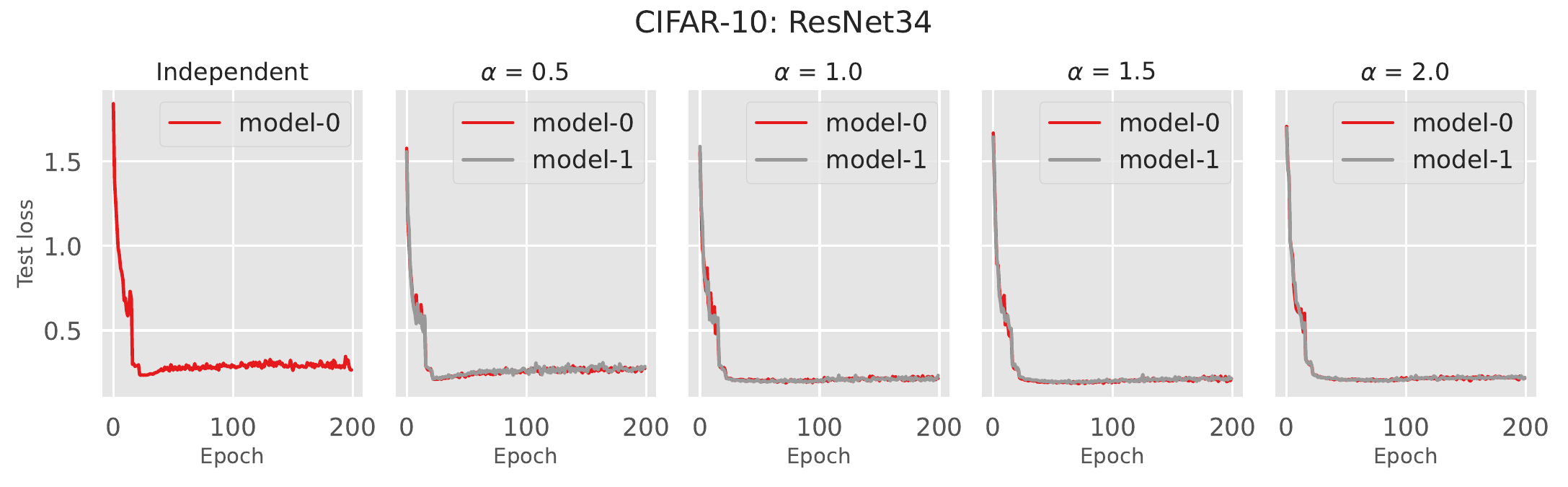}
}
\\
\subfloat{
\includegraphics[width=0.75\textwidth]{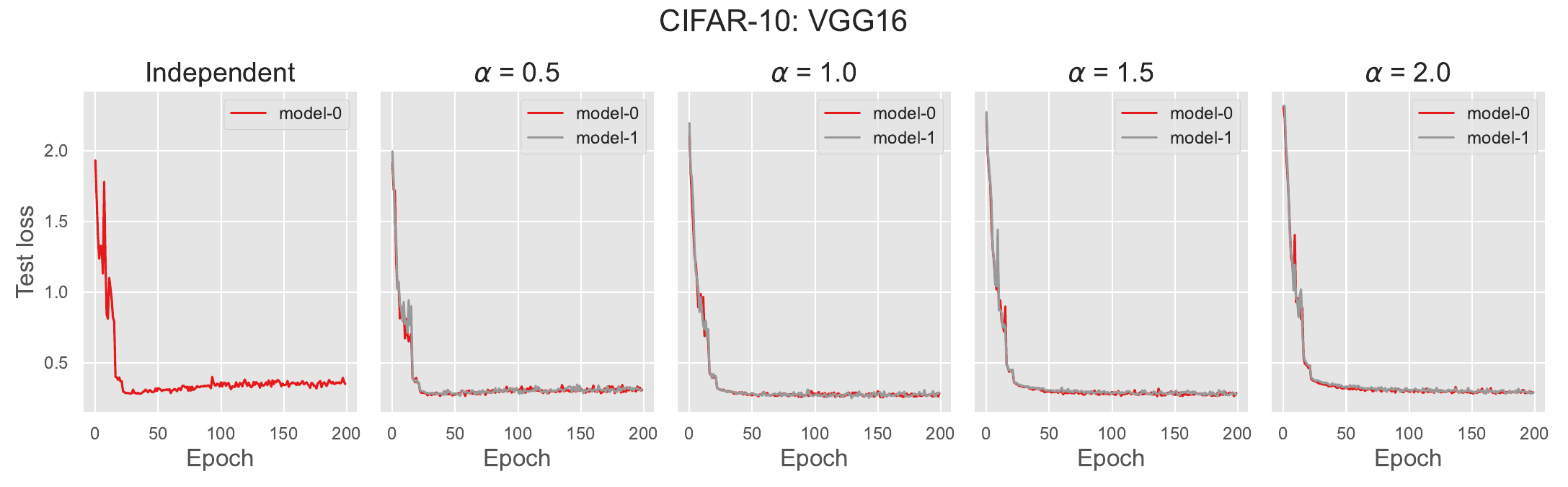}
}
\caption{Test loss for the \texttt{CIFAR10} dataset}
\end{figure*}

\begin{figure*}[h]
\centering
\subfloat{
\includegraphics[width=0.75\textwidth]{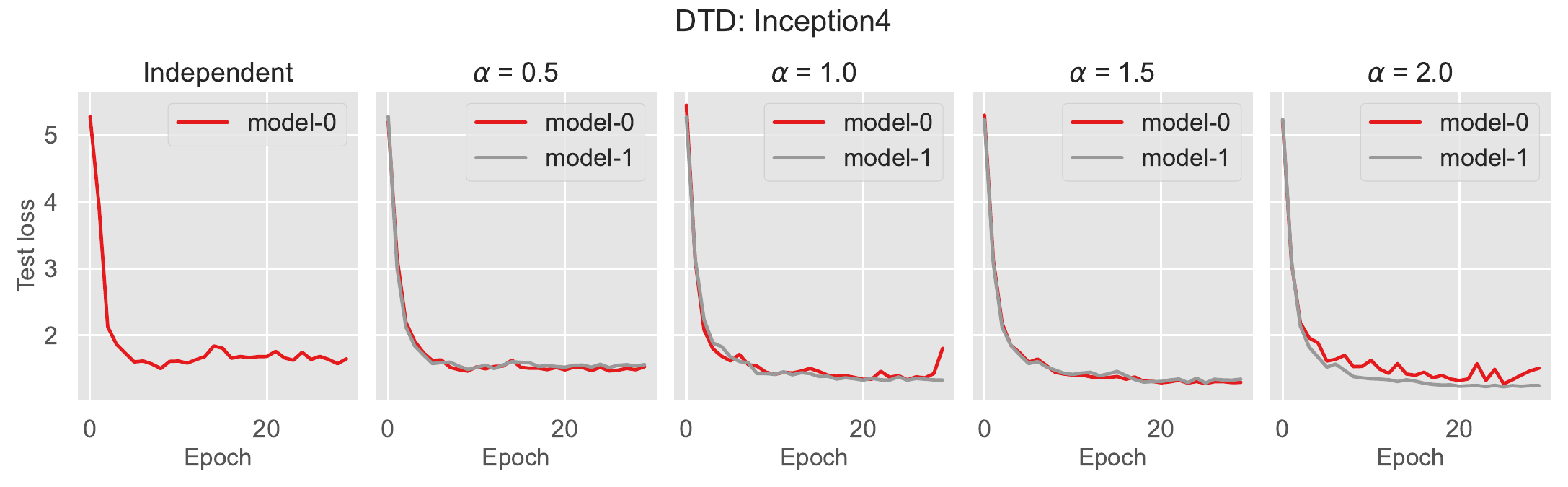}
}
\\
\subfloat{
\includegraphics[width=0.75\textwidth]{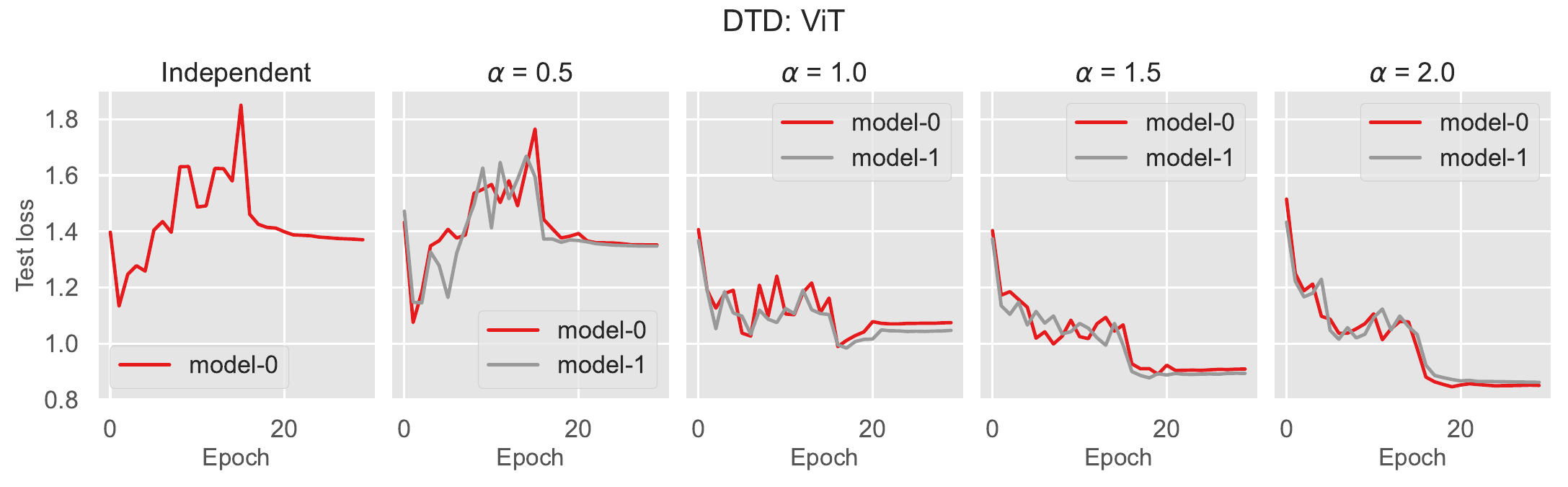}
}
\\
\subfloat{
\includegraphics[width=0.75\textwidth]{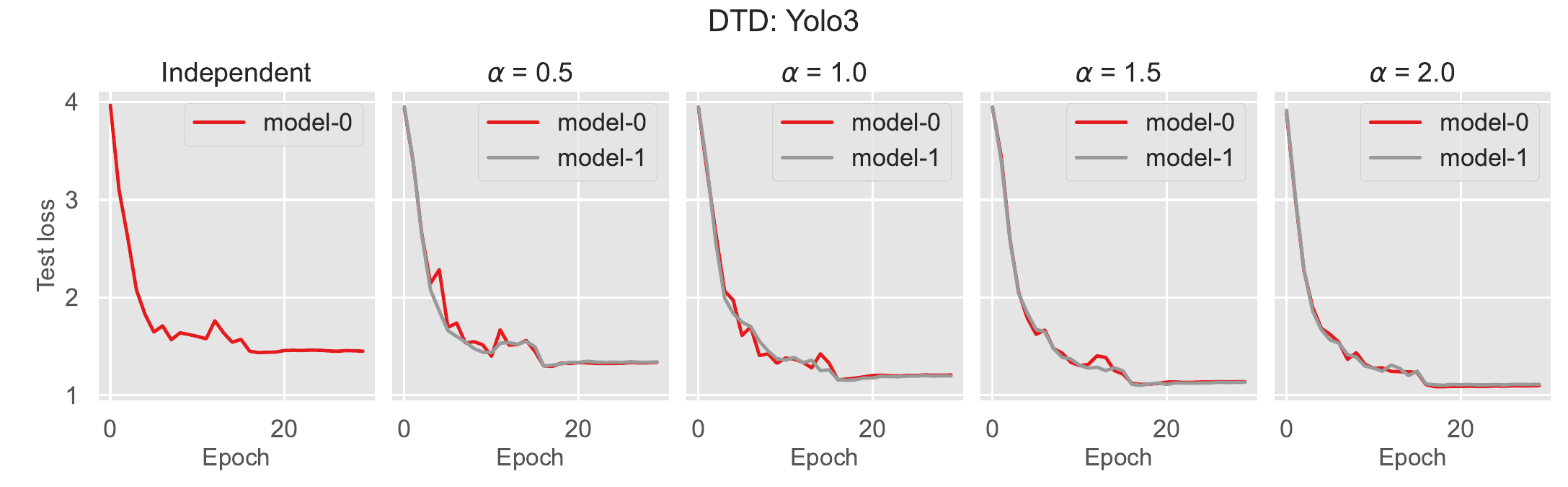}
}
\caption{Test loss for the \texttt{DTD} dataset}
\end{figure*}